\documentclass{article}
\usepackage[T1]{fontenc}
\usepackage{graphicx} 
\usepackage{natbib}
\usepackage{float}
\usepackage{amsmath}
\usepackage{amssymb}
\usepackage[a4paper, total={6in, 8in}]{geometry}
\usepackage{subcaption}
\usepackage[inline]{enumitem}
\usepackage{multirow}
\usepackage{url}
\usepackage{subcaption}
\usepackage{booktabs}
\usepackage{threeparttable}
\usepackage{xcolor}

\makeatletter
\newcommand\blfootnote[1]{%
  \begingroup
  \renewcommand{\@makefntext}[1]{\noindent\makebox[1.8em][r]#1}
  \renewcommand\thefootnote{}\footnote{#1}%
  \addtocounter{footnote}{-1}%
  \endgroup
}
\makeatother

\title{Evaluating Deep Learning Models for Fault Diagnosis of a Rotating Machinery with Epistemic and Aleatoric Uncertainty}
\author{\it Reza Jalayer$^1$, Masoud Jalayer$^2$, Andrea Mor$^{3*}$, Carlotta Orsenigo$^3$, Carlo Vercellis$^3$\blfootnote{ORCID. Reza Jalayer: 0000-0003-3440-5658. Masoud Jalayer: 0000-0001-8013-8613. Andrea Mor: 0000-0001-6131-7229. Carlotta Orsenigo: 0000-0001-8688-414X. Carlo Vercellis: 0000-0002-5020-3688.} \\
{\small \it $^1$Faculty of Engineering and Natural Sciences}\\
{\small \it Tampere University, Finland}\\
{\small reza.jalayer@tuni.fi}\\
{\small \it $^2$Department of Information and Communications Engineering}\\
{\small \it Aalto University, Finland}\\
{\small masoud.jalayer@aalto.fi}\\
{\small \it $^3$Department of Management, Economics and Industrial Engineering}\\
{\small \it Politecnico di Milano, Italy}\\
{\small \{andrea.mor,carlotta.orsenigo,carlo.vercellis\}@polimi.it}\\
{\small $^*$Corresponding author}\\
}
\date{}

\begin{document}

\maketitle

\section*{Abstract}
Uncertainty-aware deep learning models allow the reliable detection of faults when out-of-distribution (OOD) data arise from unseen faults (epistemic uncertainty) or the presence of noise (aleatoric uncertainty).
In this paper, we present the first comprehensive comparative study of state-of-the-art uncertainty-aware DL architectures for fault diagnosis in rotating machinery, where different scenarios affected by epistemic uncertainty and different types of aleatoric uncertainty are investigated.
To distinguish between in-distribution and OOD data, two uncertainty thresholds, one introduced in this paper, are alternatively applied.
Our empirical findings offer guidance to practitioners and researchers who have to deploy real-world uncertainty-aware fault diagnosis systems. In particular, in the presence of epistemic uncertainty, all tested models are capable of effectively detecting, on average, a substantial portion of OOD data across all the scenarios. However, deep ensemble models show superior performance, independently of the uncertainty threshold used for discrimination.
In the presence of aleatoric uncertainty, the noise level plays an important role. Specifically, low noise levels hinder the models' ability to effectively detect OOD data. Even in this case, deep ensemble models exhibit a milder degradation in performance, dominating the others. These achievements, combined with their shorter inference time, make deep ensemble architectures the preferred choice.

\noindent \textbf{Keywords:}
Deep Learning, Fault Diagnosis, Industry 4.0, Out of Distribution Data, Uncertainty

\section{Introduction}
\label{sec:introduction}
Rotating machines, such as motors, turbines, and generators, are indispensable components in multiple sectors, including power generation, manufacturing, and transportation.
They are crucial for the efficient operation of these industries, providing the required mechanical power and enabling the transformation of energy from one form to another.
However, like any complex equipment, rotating machines are prone to faults and failures, which can lead to significant disruptions, financial losses, and safety hazards. 
For this reason, the development and implementation of effective and intelligent fault diagnosis techniques attracted considerable research interest in recent years~\cite{lei2020applications}.

In the context of rotating machines, deep learning (DL)-based fault diagnosis plays a prominent role, due to the ability of DL models to extract intricate patterns and features from extensive datasets~\cite{liu2018artificial,zhao2020intelligent}. 
In fact, unlike conventional machine learning methods that rely heavily on hand-crafted feature engineering, DL algorithms can automatically learn and uncover concealed fault indicators~\cite{zhang2020deep,cheng2021intelligent}.
This characteristic enables the detection and classification of faults with improved accuracy and reliability~\cite{jalayer2021fault}.
By harnessing the power of DL, fault diagnosis becomes more effective, thereby improving the overall performance of the rotating machinery~\cite{hoang2019survey}.
This improvement contributes to increased productivity, leads to substantial cost savings in industrial operations~\cite{shojaeinasab2022intelligent}, and paves the way for transformative advances in the field.

With their effectiveness proven, most DL-based methods proposed in the literature are applied under the assumption that the data available for training and the data collected for future predictions share the same distribution.
This assumption, however, might be flawed in practice, as future examples could deviate from the distribution of the data used for training.
These examples are usually referred to as out-of-distribution (OOD) data~\cite{ovadia2019can}. 
The discrepancy between historic (in-distribution) and future potential OOD data introduces uncertainty in the prediction process led by the DL models, thereby affecting their fault detection accuracy.
This uncertainty can be categorized into two main types, i.e., epistemic, often called knowledge uncertainty~\cite{abdar2021review}, and aleatoric, also referred to as data uncertainty.
Epistemic uncertainty stems from the lack of knowledge and often affects DL models trained with limited data~\cite{li2020bayesian}, such as data that do not cover exhaustively all the classes the examples might belong to. 
This results in a classification model that over-misclassifies data, as it is unaware of all the possible classes. Aleatoric uncertainty, on the other hand, is due to the inherent variability in the data, which is often caused by uncontrollable factors like noise or measurement error.
Because of their nature, all the aforementioned cases occur with high frequency. Consequently, it is relevant to develop DL algorithms that are aware of the potential uncertainty in the prediction task, and are therefore capable of discriminating between in-distribution (ID) and OOD data.

To deal with epistemic and aleatoric uncertainty, some DL architectures have been proposed.
Traditional methods, such as Monte Carlo (MC) dropout, are based on sampling to drop nodes, or weights, in the neural network while predicting the class of future data.
By performing multiple predictions for the same example, sampling-based approaches generate a distribution of the output over the known classes, instead of a single outcome typical of deterministic models.
This distribution is then exploited to quantify the degree of uncertainty affecting the classification task~\cite{gal2016dropout,mcclure2016representing,amini2018spatial,zhou2023trustworthy}.
More recent approaches are based either on Bayesian neural networks (BNNs) or on ensemble DL techniques~\cite{abdar2021review}, called deep ensemble.
In particular, BNNs recently received great interest for industrial maintenance applications, where they have been successfully adopted to address uncertainty in the prognosis of the remaining useful life~\cite{kim2020bayesian,li2020bayesian,mazaev2021bayesian,caceres2021probabilistic} and in the condition monitoring of fault diagnosis~\cite{sun2020fault,sajedi2021uncertainty,qi2022combinatorial,zhou2023uncertainty,moradi2022integration,zhou2022towards,wu2020detecting,xiao2023towards,feng2024integrating}.
BNNs rely on a distribution of the neural network weights instead of a set of unique values. 
As a result, repeated predictions for the same example generate different outcomes, giving rise to a distribution over the classes.
Similar to the previous approach, this distribution is used to compute an estimate of the underlying uncertainty ~\cite{jospin2022hands,swiatkowski2020k,gal2015bayesian,blei2017variational,salimans2015markov}.
Finally, deep ensemble methods exploit multiple deep neural networks which are initialized and trained in parallel~\cite{lakshminarayanan2017simple,hu2019mbpep,jain2020maximizing}.
In this case, the distribution of the predicted class values, used to evaluate the level of uncertainty, is obtained by collecting the outcomes generated for a given example by all the models~\cite {abdar2021review,zhang2020mix}.
Deep ensemble approaches have been recently applied in a few research studies on uncertainty-informed fault diagnosis to avoid untrustworthy decisions~\cite{tuyet2021deep,han2022out,kafunah2023uncertainty,zhang2024trustworthy}. 

While uncertainty-aware DL techniques offer the advantage of quantifying uncertainty, they are not free from challenges, such as the high computational cost of the multiple runs required to obtain the distribution over the classes for each single example~\cite{abdar2021review,zhou2023trustworthy,zhou2023uncertainty,zhou2022towards,mae2021uncertainty}.
For this reason, comparing alternative methods to derive useful insights about their effectiveness in tackling uncertainty under different conditions, which is the core of the present study, deserves much attention.

To offer an overview of prior related research and highlight the novel contributions here proposed, a review of the studies focused on uncertainty-aware DL methods in the fault diagnosis domain has been conducted.
These are listed in Table~\ref{tab:1} and are categorized according to the type of uncertainty that was explored.
As one may observe, all works are fairly recent (from 2020 onwards), demonstrating the innovativeness of the present research line in the domain at hand. 
Moreover, most of the literature focused on epistemic uncertainty, leaving aleatoric uncertainty a relatively under-explored area.
In particular, epistemic uncertainty was usually analyzed by treating fault-known classes as unknown, while aleatoric uncertainty was investigated by accounting for sensors' noise, mainly limited to Gaussian noise.
Finally, existing research predominantly exploited one method at a time for fault diagnosis, disregarding the potentiality of comparing different approaches. 

In light of this review, we aimed at enriching previous results in the field by extending the analysis in two main directions. 
On one side, we broadened the study on aleatoric uncertainty by investigating the presence of both Gaussian and non-Gaussian sensor noises.
Since these latter are commonly found in sensor measurements for fault diagnosis~\cite{guo2020enhanced,jiang2020novel}, their inclusion allows for a better alignment between research and real-world application settings.
On the other, we implemented a variety of state-of-the-art DL approaches, namely sampling by dropout, BNN, and deep ensemble, and tested their effectiveness in addressing epistemic and aleatoric uncertainty for a fault prediction task.
Therefore, our work takes the form of the first comprehensive comparative study on cutting-edge uncertainty-aware DL methods for fault diagnosis in rotating machines, where both epistemic uncertainty and different variants of aleatoric uncertainty are investigated.
Our final goal is to identify, if it exists, the dominant method among those tested in our experiments, so as to provide practical guidelines for implementing effective fault diagnosis systems in real-world scenarios.

\begin{table}[ht]
\centering
\setlength{\tabcolsep}{3pt}
\renewcommand{\arraystretch}{1.5}
\begin{tabular}{|p{9cm}|c|c|}
\hline
\multirow{2}{*}{\textbf{Reference}} & \multicolumn{2}{|c|}{\textbf{Uncertainty Type}} \\ 
\cline{2-3}
 & \textbf{Epistemic} & \textbf{Aleatoric} \\ 
\hline
\cite{wu2020detecting,tuyet2021deep,han2022out,chen2022open,zhou2023trustworthy,yi2023uncertainty,zhang2024trustworthy,yao2024uncertainty,feng2024integrating} & $\checkmark$ & $\Box$ \\
\cite{zhou2022towards,zhou2023uncertainty,kafunah2023uncertainty,xiao2023towards} & $\checkmark$ & $\checkmark$ \\
\hline
\end{tabular}
\caption{\textbf{Classification of relevant literature in the fault diagnosis domain based on the types of uncertainty addressed by DL models: epistemic uncertainty and aleatoric uncertainty.}}
\label{tab:1}
\end{table}

The paper is organized as follows. 
Section~\ref{sec:preliminaries} provides some preliminaries, including the distinction between epistemic and aleatoric uncertainty and the description of the DL approaches investigated in our study.
Section~\ref{sec:exp_analysis} illustrates the dataset used for experiments, provides details on the architecture adopted for each DL approach, and describes the uncertainty-analysis pipeline. 
Computational results are presented in Section~\ref{sec:results}, whereas conclusions and future research directions are discussed in Section~\ref{sec:conclusions}.

\section{Preliminaries}
\label{sec:preliminaries}
In this section an overview of out-of-distribution (OOD) data that can affect classification tasks is provided.
Then, three different DL architectures commonly used to cope with OOD data, i.e., sampling by dropout, Bayesian neural networks, and deep ensemble, are explained.

\subsection{Out-of-distribution data}
As mentioned in the introduction, out-of-distribution data originate from the presence of two types of uncertainties: epistemic and aleatoric.

\begin{figure}[ht]
 \centering
 \includegraphics[width=0.7\linewidth]{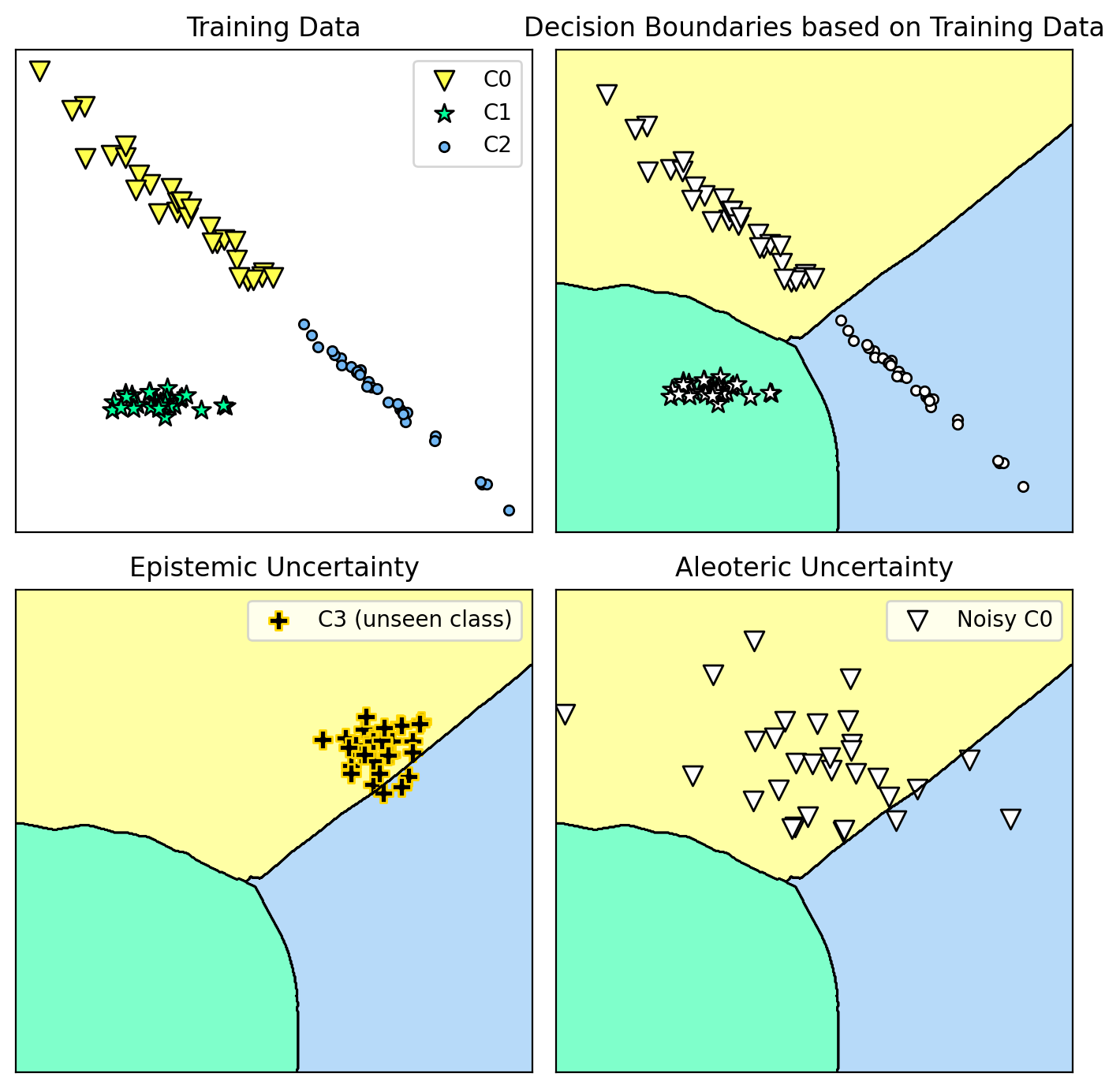}
 \caption{\textbf{Representation of epistemic and aleatoric uncertainty for a classification problem.}}
 \label{fig:1}
\end{figure}

To picture epistemic uncertainty consider a classical, uncertainty-unaware, DL model trained to predict three types of health conditions, i.e., C0, C1, and C2, in Figure~\ref{fig:1}, top left.
Such a model would not be able to properly predict the status of future examples of a new fault type (e.g., C3, Figure~\ref{fig:1}, bottom left), never observed before.
In this case the model would assign one of the known, pre-ingested classes, generating two alternative undesirable outcomes: either it would predict a wrong failure type or the healthy condition, which is possibly even worse.
Aleatoric uncertainty, instead, can occur in fault diagnosis as the result of noise affecting the sensor capturing the data from the machine during the diagnosis process.
This noise can originate from internal factors, like the sensor and its circuit elements, or external interference, such as human operator interventions or possible changes that might occur in the environment where the industrial equipment operates~\cite{li2020recent}.
In this case, the traditional DL algorithm would struggle to discern features from data surrounded by noise (Figure~\ref{fig:1}, bottom right), especially for noise above a certain threshold~\cite{qiao2020deep}, and due to the inadequate feature extraction it would mis-predict the faulty class.

The inability of quantifying uncertainty makes fault diagnosis assessment unreliable. 
This results in a poor evaluation of the industrial equipment's health, thereby increasing the risk of losing valuable resources, causing interruptions in strategic systems, or even endangering safety.
Beyond the inevitable financial costs, unreliable fault diagnosis can lead to severe consequences in critical applications, such as for nuclear power plants~\cite{cheng2023three,yao2024uncertainty}.
Epistemic and aleatoric uncertainty can be tackled by gathering additional data containing previously unseen faults, discarding noisy data or tackling the sources of noise.
In any case, however, a prediction model capable of addressing the uncertainty in the diagnostic process is required to apply the appropriate corrective actions. 

\subsection{Uncertainty-aware deep learning models}
\label{sec:ua_dlmodels}
In uncertainty-aware DL models a distribution of the predictions over the classes is used to measure the degree of uncertainty affecting the classification task.
This distribution is generated differently according to the DL technique used, as described in the following sections.

\subsubsection{Sampling by dropout}
\label{sec:sample_drop}

Dropout was introduced in DL architectures as a regularization method to mitigate overfitting.
In uncertainty-unaware scenarios, it randomly deactivates a sample of nodes, or weights, of the neural network during the training phase~\cite{srivastava2014dropout}, where the sampling is performed according to a Monte Carlo approach (MC dropout)~\cite{abdar2021review}.
In the presence of uncertainty, dropout is instead applied at inference time.
Specifically, $K$ dropout samplings are applied on the trained neural network and the resulting models are used to generate a distribution of the predictions over the classes for each example to classify.
This distribution is later exploited to estimate the level of uncertainty of the neural network on the given set of data, as described in Section~\ref{sec:exp_analysis}.

\subsubsection{Bayesian neural network}
 In a classical neural network the weights are deterministic, leading to a deterministic mapping between inputs and outputs.
 In contrast, BNNs introduce stochasticity by representing the weights as random variables following a given distribution.
This is known as the posterior distribution and is learned during the training phase.
The distribution of the outcomes over the classes for a given example is, in this case, achieved through $K$ predictions based on $K$ different samples of the weights from the posterior distribution.

\subsubsection{Deep ensemble}

In a deep ensemble architecture several base learners, given by DL algorithms, are trained independently. 
Instead of having a point estimate output from a single DL model, the set of outcomes produced by the base learners is leveraged in order to generate the distribution over the classes for each example to predict. 
The base learners can be identical~\cite{tuyet2021deep} or may vary in structure.
In particular, their diversity has proven to be useful for achieving more accurate quantification of uncertainty in fault diagnosis \cite{han2022out}.

\section{Experimental settings}
\label{sec:exp_analysis}

In this section we first illustrate the dataset used for the experiments. Then, we present the uncertainty-aware DL architectures investigated in our study, and outline the process for generating out-of-distribution data from both the epistemic and the aleatoric uncertainty perspective. Finally, we explain how uncertainty is measured and describe the uncertainty analysis pipeline.

\subsection{Datasets and data preparation}
\label{subsec:cwru}
\begin{figure}[ht]
 \centering
 \includegraphics[width=8cm]{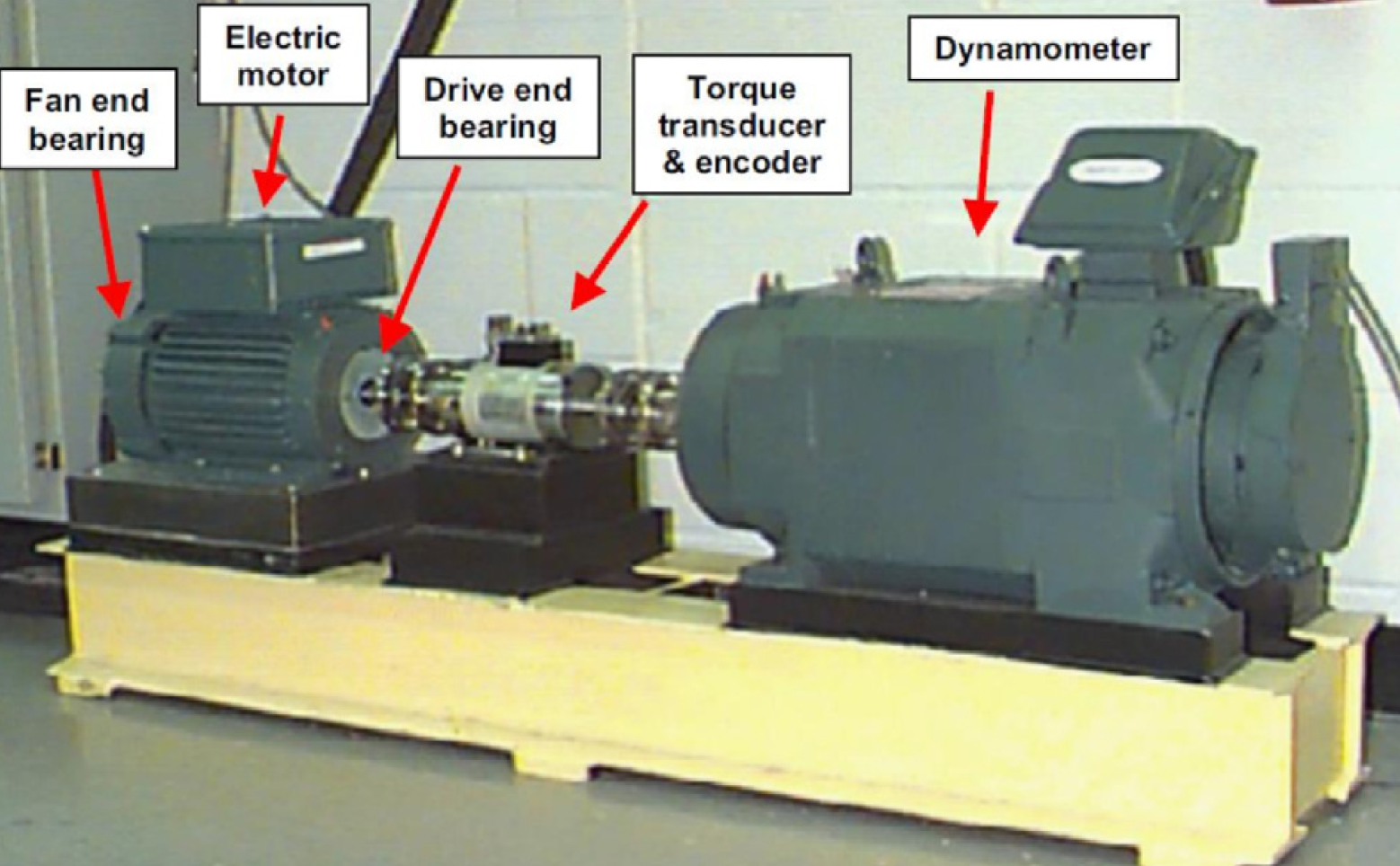}
 \caption{\textbf{The setup for the CWRU dataset from left to right: fan-end bearing, 2 HP electric motor, drive-end bearing, torque transducer and encoder, and dynamometer. This picture is taken from the CWRU website.}}
 \label{fig:6}
\end{figure}

The Case Western Reserve University (CWRU) dataset\footnote[1]{\url{https://engineering.case.edu/bearingdatacenter}} is an open-access dataset extensively analyzed in fault diagnosis studies, and is a standard reference for evaluating DL algorithms~\cite{neupane2020bearing}.
This dataset gathers information from a bearing testing platform consisting of two bearings located at the fan-end and drive-end sections, a 2 HP electric motor, a torque transducer and encoder, and a dynamometer, as represented in Figure~\ref{fig:6}.
In particular, the dataset collects data on the operational states of the rotating machine, which are categorized into either healthy (normal) or three distinct faulty conditions (i.e., ball-bearing, inner-race, and outer-race), depending on the fault's location.
Faults for the outer-race are further divided into centered, orthogonal, and opposite, based on the load zone. 
This arrangement leads to a final set of six different labels to predict, as indicated in Table~\ref{tab:2}.

From the set of available examples, we focused on the data recorded at a sampling rate of 12000 samples per second (``12k drive-end fault'' data), which contains the healthy and faulty conditions mentioned above.

\begin{table}[ht]
\centering
\setlength{\tabcolsep}{3pt}
\renewcommand{\arraystretch}{1.3}
\begin{tabular}{|l|c|}
\hline
\textbf{Fault Type} & \textbf{Label} \\
\hline
Healthy Condition (Normal) & 0 \\
Inner-race fault & 1 \\
Ball-bearing fault & 2 \\
Outer-race fault – Centered & 3 \\
Outer-race fault – Orthogonal & 4 \\
Outer-race fault – Opposite & 5 \\
\hline
\end{tabular}
\caption{\textbf{Labels in the CWRU dataset, representing either the healthy (0) or a faulty condition (1–5).}}
\label{tab:2}
\end{table}

For each bearing condition the raw data consists of a time series, which was divided into a set of bursts of fixed (512 data points) length.
These were extracted sequentially along the time series, with a forward shift of 200 data points.
To obtain a balanced dataset for each bearing condition we extracted the same number (2,624) of bursts from each time series.

\begin{figure}[ht]
 \centering
 \includegraphics[width=\linewidth]{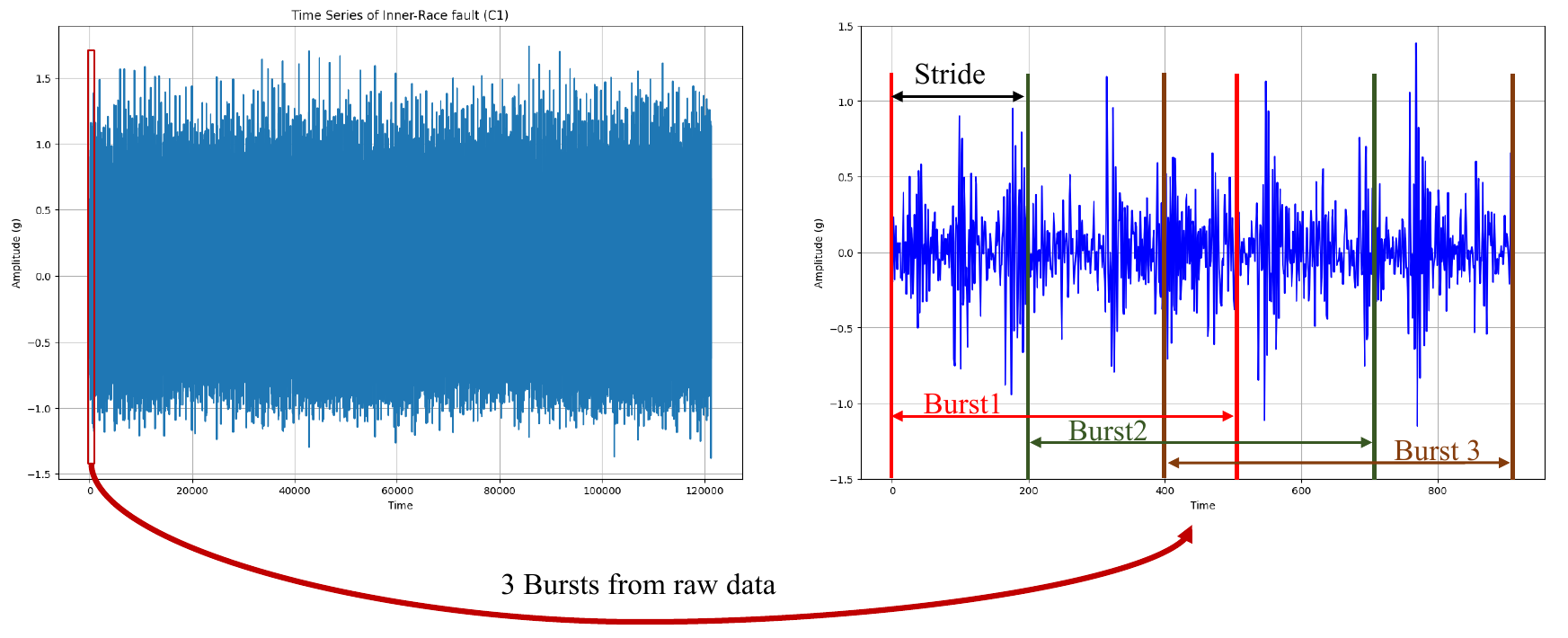}
 \caption{\textbf{Extraction of the first three bursts from a raw time series.}}
 \label{fig:7}
\end{figure}

At the end of this process, depicted in Figure~\ref{fig:7}, we achieved a final dataset comprising 15,744 examples (bursts). Of these, 70\% was randomly selected and used for training. The 30\% portion was then further divided, assigning 70\% to the test set and 30\% to the validation set\footnote[2]{The code for the data generation and the fault diagnosis tasks will be made publicly available after publication.}. With respect to ID and OOD data, the training is composed by ID data only, whereas validation and test are composed by both ID and OOD data. The details of the ID/OOD data distribution is described in detail in Section \ref{subsec:OOD GENERATION}.

\subsection{Uncertainty-aware DL architectures}
\label{subsec:unc_architectures}
In what follows we describe the architectures used in the study for each of the methods introduced in Section~\ref{sec:preliminaries}, i.e., sampling by dropout, Bayesian neural network, and deep ensemble. All architectures share a common element represented by a core building block defined by a 1D convolutional layer (Conv1D) followed by Max-Pooling.
This block will be hereafter denoted as Conv+Pool.
Moreover, all architectures employ an input layer comprising 512 nodes, matching the length of each input example (burst).
The list of hyperparameters used in the experiments for each architecture is provided in Appendix~\ref{app:A}.

\subsubsection{Sampling by dropout}

\begin{figure}[ht]
 \centering
 \includegraphics[width=\linewidth]{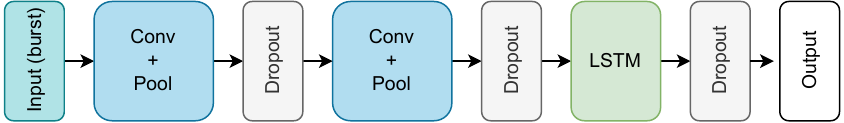}
 \caption{\textbf{Convolutional LSTM with Bernoulli dropout.}}
 \label{fig:9}
\end{figure}

In this architecture a Convolutional LSTM with Bernoulli dropout is employed, as illustrated in Figure~\ref{fig:9}, where the input layer comprises 512 nodes so as to match the length of each example (burst).
In particular, a Conv+Pool block is replicated twice and is followed by an LSTM layer whose output is fed to a dropout layer.
Finally, a dense softmax layer is used to produce 
a prediction, consisting of a vector where each component reports, for a given example, the likelihood of being in that class.
This architecture was chosen since it turned out to be the most effective for addressing the same prediction task in the absence of uncertainty~\cite{jalayer2021fault}. In this case, instead, it can be exploited to quantify the level of uncertainty by sampling different dropouts in a Monte Carlo manner, to generate a distribution of the predictions over the classes. Henceforth, we will refer to this architecture as ConvLSTM-D.

It is worth mentioning that, the number $K$ of predictions obtained for each example impacts the uncertainty quantification, independently of the model used for classification and, consequently, is a parameter to tune. For sampling by dropout the experiments were conducted by setting $K=10$, since this was the minimum value for which stable results were observed.

\subsubsection{Bayesian neural network}

\begin{figure}[ht]
 \centering 
 \includegraphics[width=0.7\linewidth]{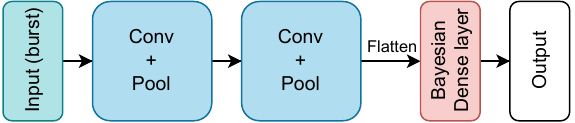}
 \caption{\textbf{Bayesian neural network architecture.}}
 \label{fig:10}
\end{figure}

The BNN architecture considered in this work is depicted in Figure~\ref{fig:10}.
Two Conv+Pool blocks are first employed. The output is then flattened and fed to the fully connected layers.
These latter layers implement the Bayesian logic in the weights of the network, which are defined as random variables following a Gaussian distribution.
It is worth mentioning that, during the training phase node-based dropout is used between the two Bayesian dense layers in order to mitigate overfitting.
As for sampling by dropout, the number $K$ of predictions generated for each example by BNN was set to 10.

\subsubsection{Deep ensemble}

\begin{figure}[ht]
 \centering
 \includegraphics[width=0.7\linewidth]{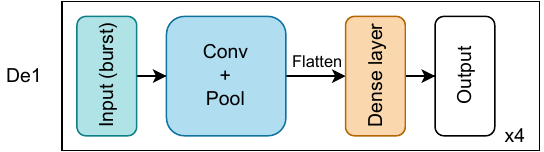}
 \caption{\textbf{Architecture of deep ensemble De1.}}
 \label{fig:11}
\end{figure}

\begin{figure}[ht]
 \centering
 \includegraphics[width=0.7\linewidth]{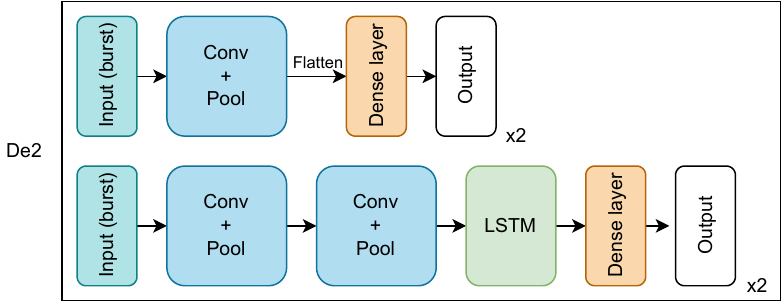}
 \caption{\textbf{Architecture of deep ensemble De2.}}
 \label{fig:12}
\end{figure}

To investigate the deep ensemble approach two configurations were explored.
The first, referred to as De1 and depicted in Figure~\ref{fig:11}, is composed of four base learners all with identical architecture. 
This is defined by an input layer followed by a Conv+Pool. Then, the resulting features are flattened to be fed to two dense layers set in sequence. 
The second configuration, referred to as De2 and shown in Figure~\ref{fig:12}, consists of four base learners with two different architectures.
Two base learners follow the same structure of the learners of De1. 
The remaining ones, instead, are built by two Conv+Pool blocks followed by an LSTM layer and two dense layer. During training, node-based dropout is applied between the two dense layers.

In deep ensemble the number of base learners comprised in the architecture automatically determines the number $K$ of predictions generated for each example.
This is why for both De1 and De2 experiments were conducted by using $K=4$.
As for the previous architectures, the number of base learners, and therefore the value of $K$, was set to the minimum value for which stable results were observed.

\subsection{OOD data generation}
\label{subsec:OOD GENERATION}

In our experimental framework training was performed on in-distribution data, whereas validation and test were carried out on a dataset comprising both ID and OOD data.
In this section, we describe how OOD were generated to alternatively test either the epistemic or the aleatoric uncertainty perspective.

To simulate epistemic uncertainty we employed a hold-out strategy by treating examples from all but one class as ID data, and using examples from the hold-out class as OOD data.
By iteratively changing the class to exclude during training, for the CWRU dataset we generated six scenarios, as exemplified by Table~\ref{tab:3}.

\begin{figure}[ht]
 \centering
 \includegraphics[width=0.9\linewidth]{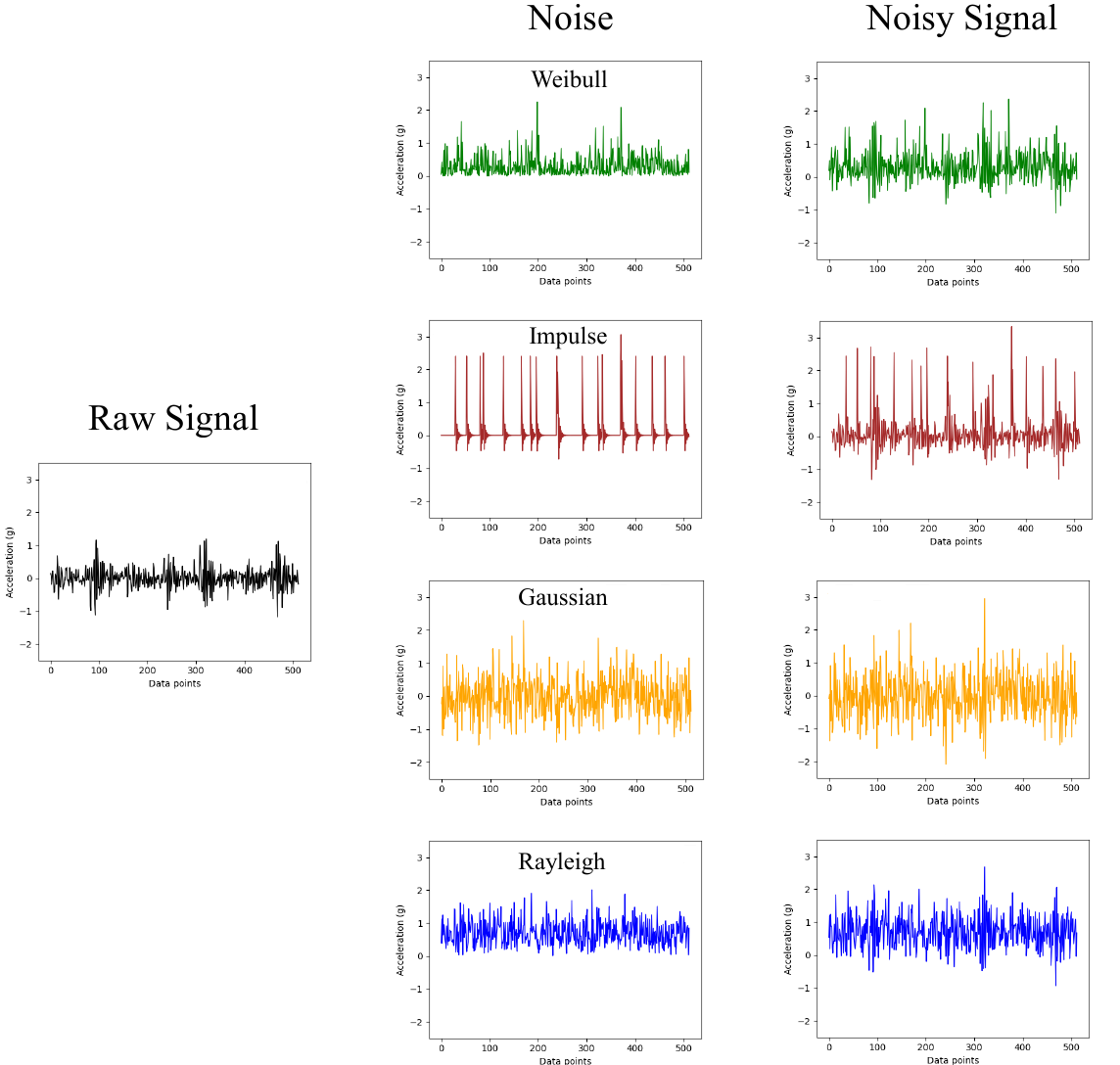}
 \caption{\textbf{The four noise types and their effect when applied to a raw signal.}} 
 \label{fig:13}
\end{figure}

To reproduce aleatoric uncertainty, instead, we generated OOD data by injecting noise into the validation and the test set. 
In particular, we applied Gaussian noise, which is usually analyzed in aleatoric uncertainty studies related to fault diagnosis, as well as three non-Gaussian noise types, i.e. Impulse, Rayleigh, and Weibull noise.
Impulse noise is frequently investigated in fault diagnosis research due to its presence during external shocks or sudden collisions in bearing operations~\cite{lu2022bearing, moshrefzadeh2019spectral,hebda2022infogram,peng2024adaptive}.
Rayleigh and Weibull noises have been explored in bearing fault diagnosis studies~ \cite{du2022fault,guo2020enhanced}.
The effect of applying the four noise types to a raw signal is shown Figure~\ref{fig:13}.

\begin{table}[ht]
\centering
\setlength{\tabcolsep}{3pt}
\renewcommand{\arraystretch}{1.3}
\begin{tabular}{|l|c|c|c|c|c|c|}
\hline
\multirow{2}{*}{\textbf{Scenario Number}} & \multicolumn{6}{|c|}{\textbf{Class Labels in Validation/Test}} \\ 
\cline{2-7}
 & \textbf{0} & \textbf{1} & \textbf{2} & \textbf{3} & \textbf{4} & \textbf{5} \\ 
\hline
No.0 & OOD & ID & ID & ID & ID & ID \\
No.1 & ID & OOD & ID & ID & ID & ID \\
No.2 & ID & ID & OOD & ID & ID & ID \\
No.3 & ID & ID & ID & OOD & ID & ID \\
No.4 & ID & ID & ID & ID & OOD & ID \\
No.5 & ID & ID & ID & ID & ID & OOD \\
\hline
\end{tabular}
\caption{\textbf{Epistemic uncertainty scenarios.}}
\label{tab:3}
\end{table}

Furthermore, to produce signals with varying noise levels we applied the aforementioned noise types with different noise levels. This was achieved by adjusting the signal-to-noise ratio (SNR), which represents the ratio between the power of the signal and the power of the noise, as follows:

\begin{equation}\label{eq:7}
SNR = 10 \log_{10} \left ( \frac{\text{Signal power}}{\text{Noise power}} \right).
\end{equation}

\subsection{Uncertainty quantification}
\label{subsec:UQ}

As described in Section \ref{sec:ua_dlmodels}, uncertainty-aware models generate, for each example, $K$ predictions each represented by a vector of scores, where each score is the likelihood of belonging to a given class.
These scores are used to quantify the level of uncertainty in the predictions of the model.
The most frequently adopted measure of uncertainty is given by the entropy defined, for a given example, as follows:
\begin{equation}\label{eq:8nt}
   H = -\sum^C_{c=1}  ((\sum^K_{k=1} \frac {p_{c}^k}{K} )  log (\sum^K_{k=1} \frac {p_{c}^k}{K} )),
\end{equation}

where $p_{c}^k$ is the likelihood of the example of being in class $c$, $c \in [1, \ldots, C]$, according to prediction $k, k \in [1,\ldots,K]$.
For a given example, a high entropy value implies a great uncertainty in its prediction and, therefore, a great likelihood of being part of OOD data.

Finding the threshold to apply on the entropy of the predictions to discriminate properly between ID and OOD data is a challenging task. Setting the threshold too high leads to a major risk of flagging OOD data as trustworthy. Conversely, if the threshold is too low the risk of misclassifying ID data increases.
The ideal value would be the one for which the number of OOD data correctly detected is maximized, and the number of ID data erroneously deemed as untrustworthy is minimized.

As the OOD examples are unknown during training, authors in~\cite{han2022out} computed the entropy threshold on the ID data used for validation by resorting to the inter-quartile range (IQR) proximity rule.
Specifically, they defined this threshold, here denoted as $\tau_1$, as follows:
\begin{equation}\label{eq:9}
\begin{split}
   \tau_1 &= Q_3 (H_{\text{ID}}) + 1.5 (Q_3(H_{\text{ID}}) - Q_1 (H_{\text{ID}})) \\
          &= Q_3 (H_{\text{ID}}) + 1.5 \text{IQR} (H_{\text{ID}}),
\end{split}
\end{equation}
where $Q_1 (H_{\text{ID}})$ and $Q_3 (H_{\text{ID}})$ are, respectively, the first and the third quartile of the entropy values computed for the ID validation set.
In what follows, we will refer to $\tau_1$ as the threshold based on the ``uncertainty outlier detection''.

In this paper, we propose a novel method for computing the entropy threshold based on the confusion matrix reported in Table~\ref{tab:14_cm}.
This matrix provides an overview of the performance of the model on the validation set once the entropy threshold is fixed. 
In particular, True Positive (TP) is the number of OOD examples labeled as untrustworthy ($H$ above, or equal to, the threshold) by the model, True Negative (TN) is the number of ID examples labeled as trustworthy ($H$ below the threshold), False Positive (FP) is the number of ID examples labeled as untrustworthy ($H$ above, or equal to, the threshold), and False Negative (FN) is the number of OOD examples labeled as trustworthy ($H$ below the threshold).
Based on this matrix, we defined the new entropy threshold as

\begin{equation} \label{eq:tau2}
\tau_2 =  \arg\max_{H \in H_{V}} f_1(H),
\end{equation}

where $H_{V}$ is the set of entropy values computed for the examples in the validation set, and $f_1(H)$ denotes the F1-score, i.e. the harmonic mean of Precision (TP/(TP+FP)) and Recall (TP/(TP+FN)), obtained by using the value $H$ as the threshold.
Hereafter, we will refer to $\tau_2$ as the threshold based on the ``uncertainty $f_1$ score''.

\begin{table}[ht]
\centering
\setlength{\tabcolsep}{3pt}
\renewcommand{\arraystretch}{1.5}
\begin{tabular}{|c|c|c|c|}
\hline
\multicolumn{2}{|c|}{\multirow{2}{*}{\begin{tabular}[c]{@{}c@{}} \textbf{Confusion Matrix of}\\ \textbf{Uncertainty Analysis} \end{tabular}}} & \multicolumn{2}{c|}{\textbf{Prediction}} \\ 
\cline{3-4}
\multicolumn{2}{|c|}{} & \textbf{Trustworthy} & \textbf{Untrustworthy} \\ 
\hline
\multirow{2}{*}{\textbf{Actual}} & \textbf{ID}  & True Negative  & False Positive \\ 
 & \textbf{OOD} & False Negative & True Positive  \\ 
\hline
\end{tabular}
\caption{\textbf{The confusion matrix of untrustworthy analysis.}}
\label{tab:14_cm}
\end{table}

Once the threshold is determined, the analysis proceeds with the classification of the test examples into trustworthy and untrustworthy data.
As for the validation phase, this is achieved by applying $\tau_1$ or $\tau_2$ to the entropy value generated, for each example, by the trained uncertainty-aware models.

\begin{figure}[H]
 \centering
 \includegraphics[width=0.9\linewidth]{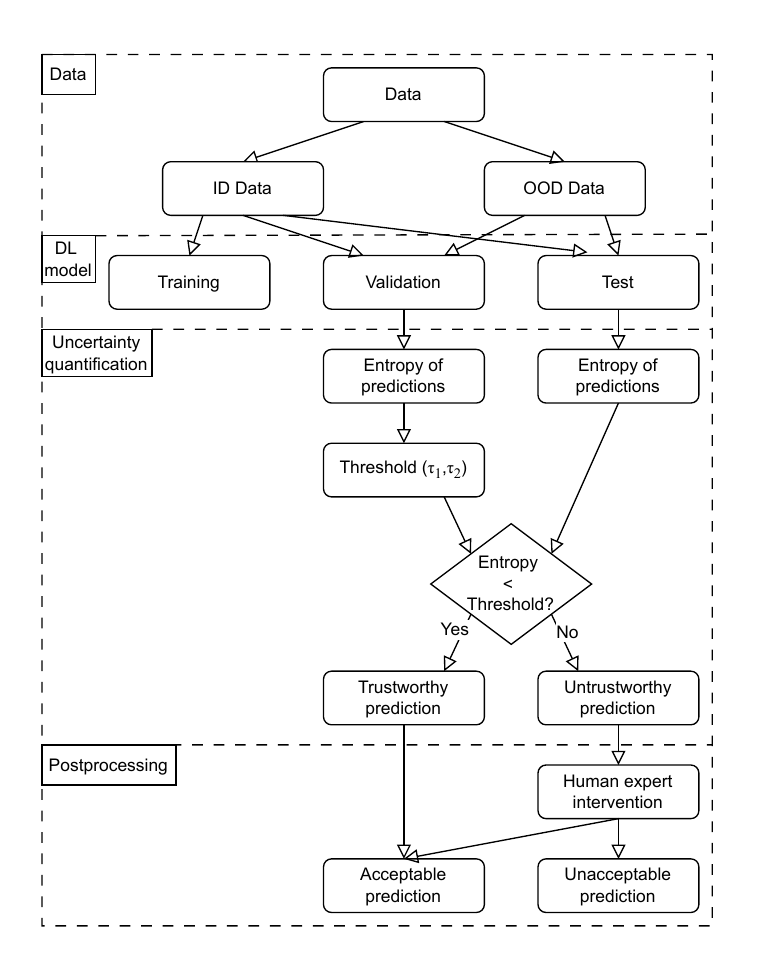}
 \caption{\textbf{Representation of the uncertainty analysis pipeline.}}
 \label{fig:15}
\end{figure}

A schematic representation of the uncertainty analysis pipeline is provided in Figure~\ref{fig:15}.
The pipeline comprises a final post-processing stage which may require the intervention of domain experts to check for the reliability of the OOD data predictions.

\section{Results \& Discussion}
\label{sec:results}

In this section we illustrate the prediction results obtained on the test examples by the architectures proposed in Section~\ref{subsec:unc_architectures}, for both epistemic and aleatoric uncertainty scenarios.
The impact of the two approaches for computing the entropy threshold, presented in Section~\ref{subsec:UQ}, is also discussed.

\subsection{Epistemic uncertainty scenarios}

\begin{figure}[H]
	\centering
	\begin{subfigure}{0.33\linewidth}
		\centering
		\includegraphics[width=\linewidth]{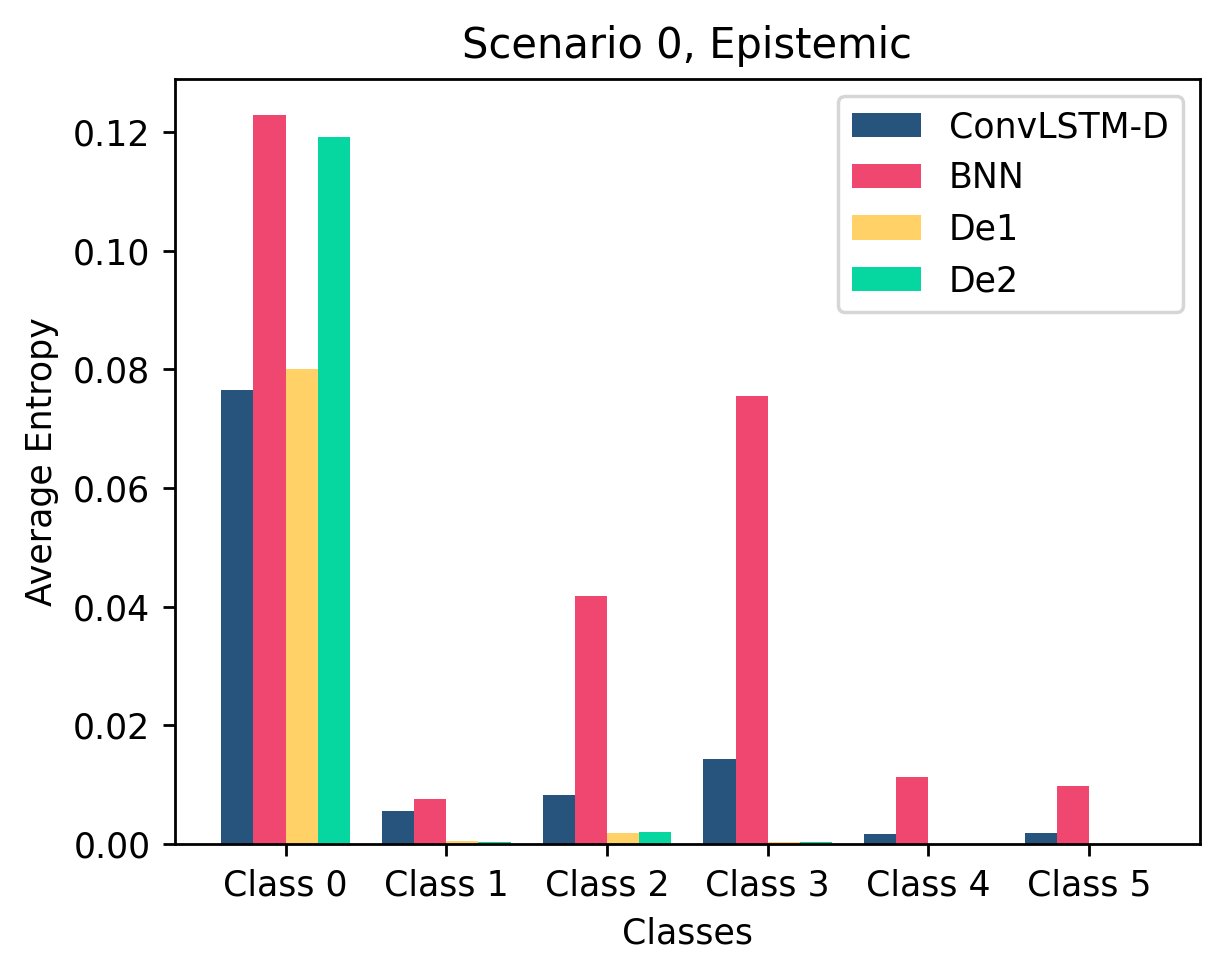}
		\caption{Scenario No.0}
	\end{subfigure}
	\begin{subfigure}{0.33\linewidth}
		\centering
		\includegraphics[width=\linewidth]{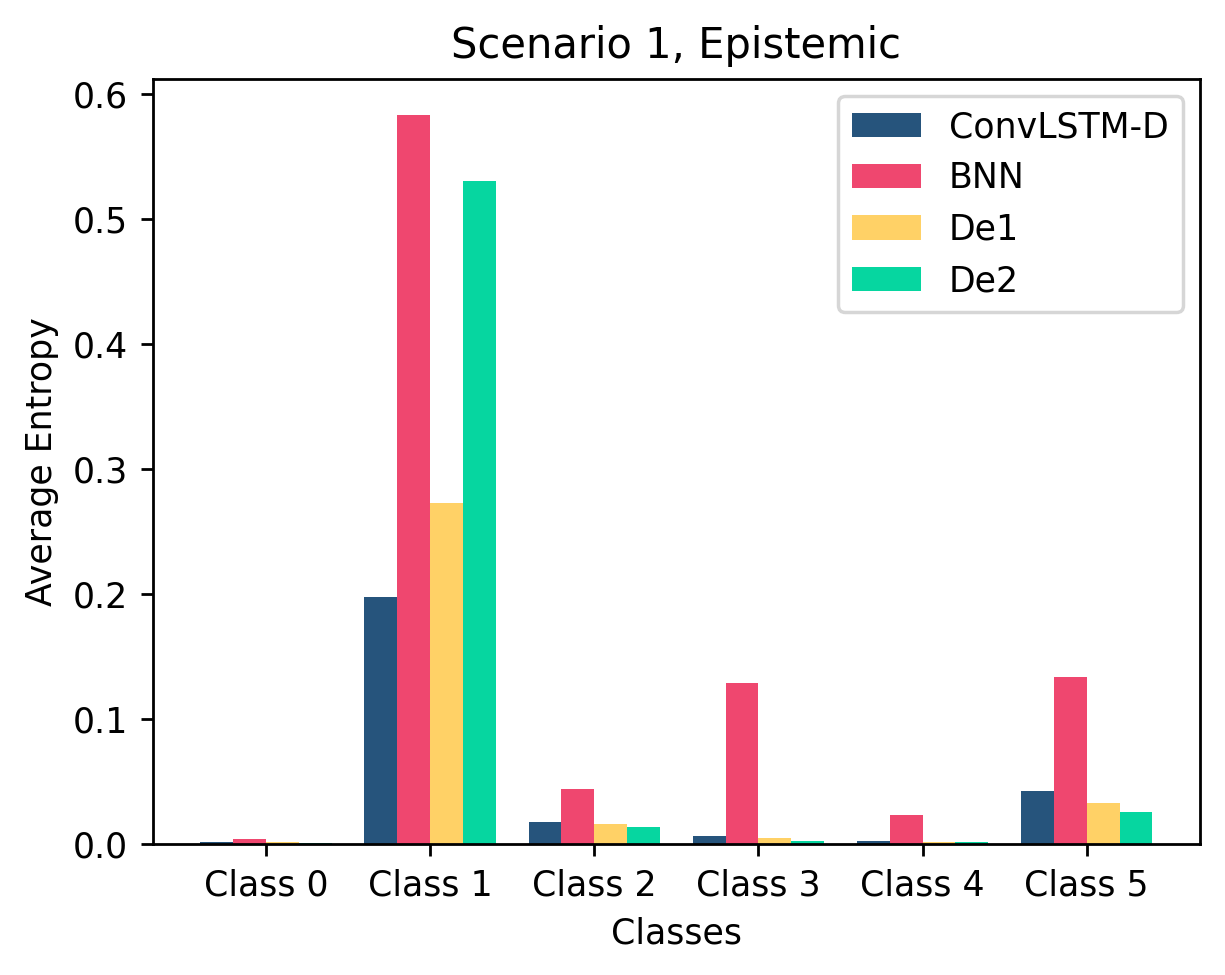}
		\caption{Scenario No.1}
	\end{subfigure}

	\vspace{2mm} 

	\begin{subfigure}{0.33\linewidth}
		\centering
		\includegraphics[width=\linewidth]{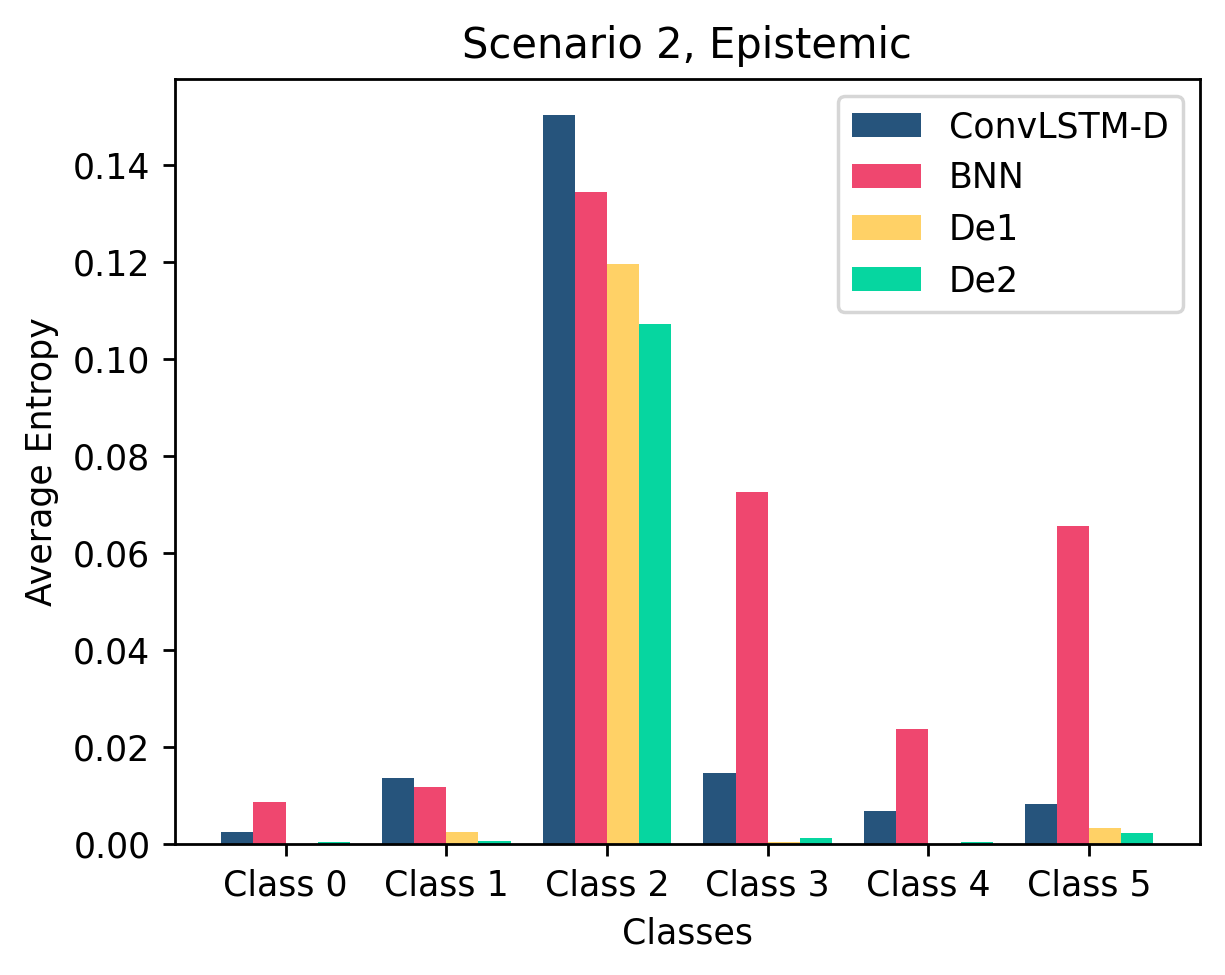}
		\caption{Scenario No.2}
	\end{subfigure}
	\begin{subfigure}{0.33\linewidth}
		\centering
		\includegraphics[width=\linewidth]{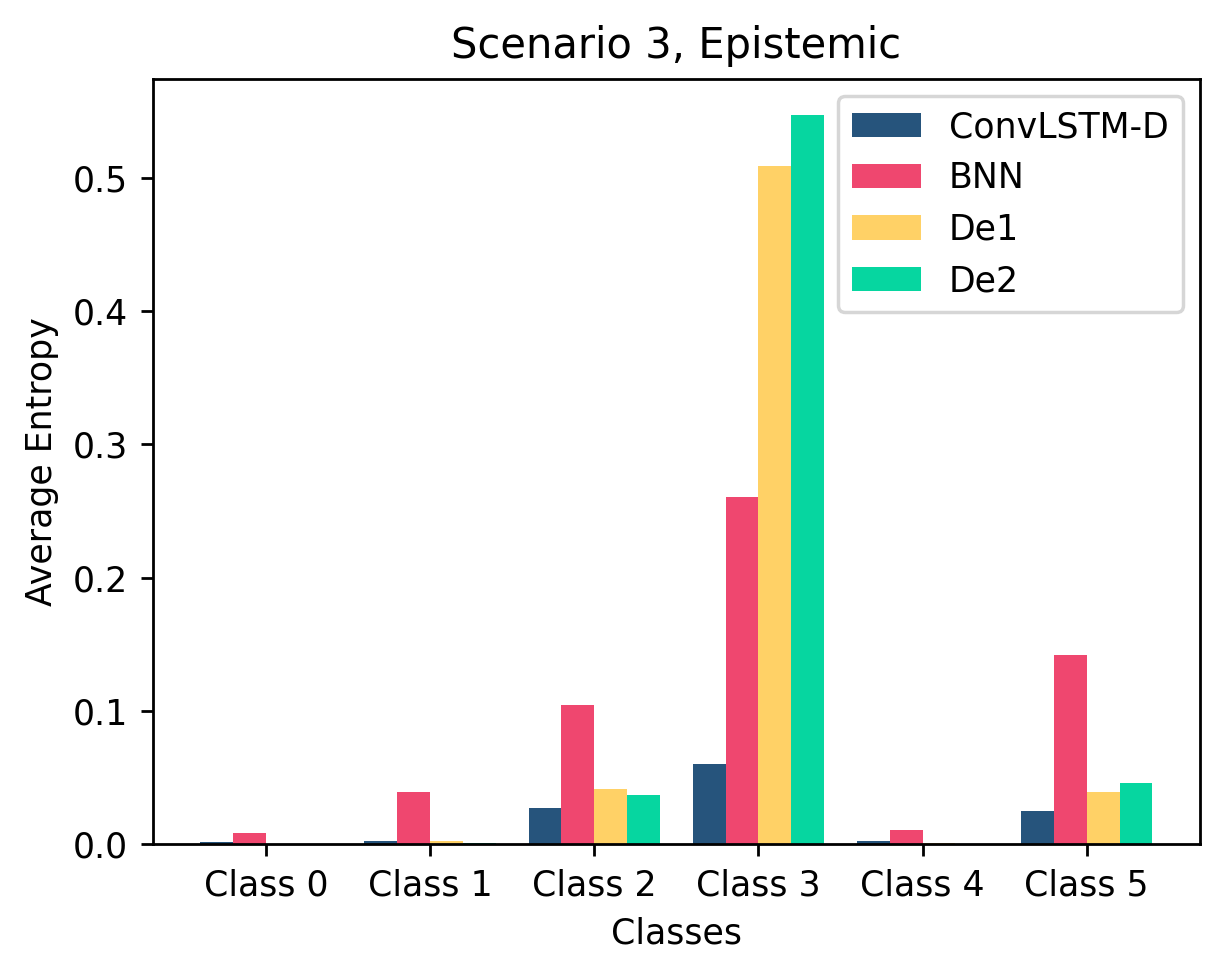}
		\caption{Scenario No.3}
	\end{subfigure}

	\vspace{2mm}

	\begin{subfigure}{0.33\linewidth}
		\centering
		\includegraphics[width=\linewidth]{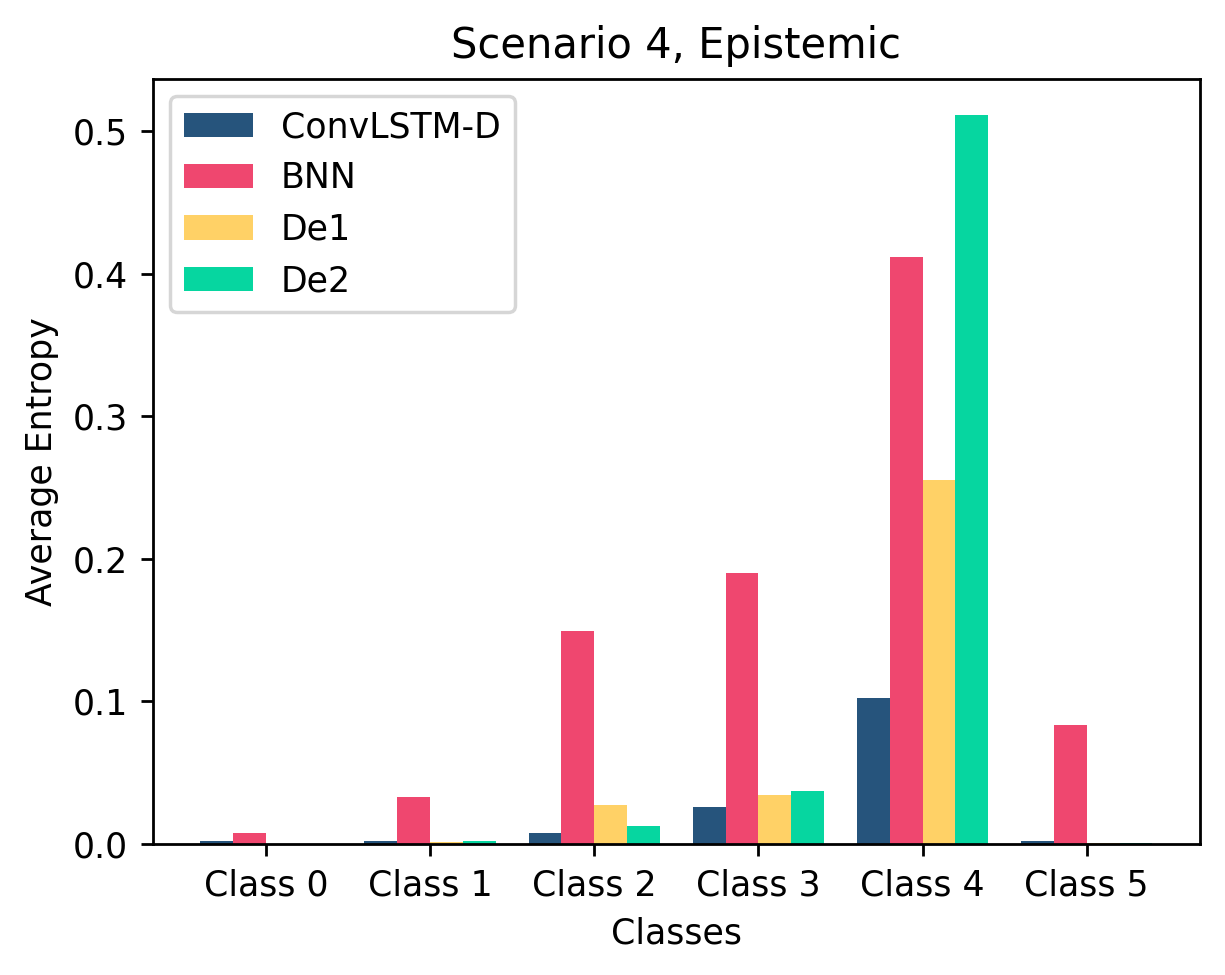}
		\caption{Scenario No.4}
	\end{subfigure}
	\begin{subfigure}{0.33\linewidth}
		\centering
		\includegraphics[width=\linewidth]{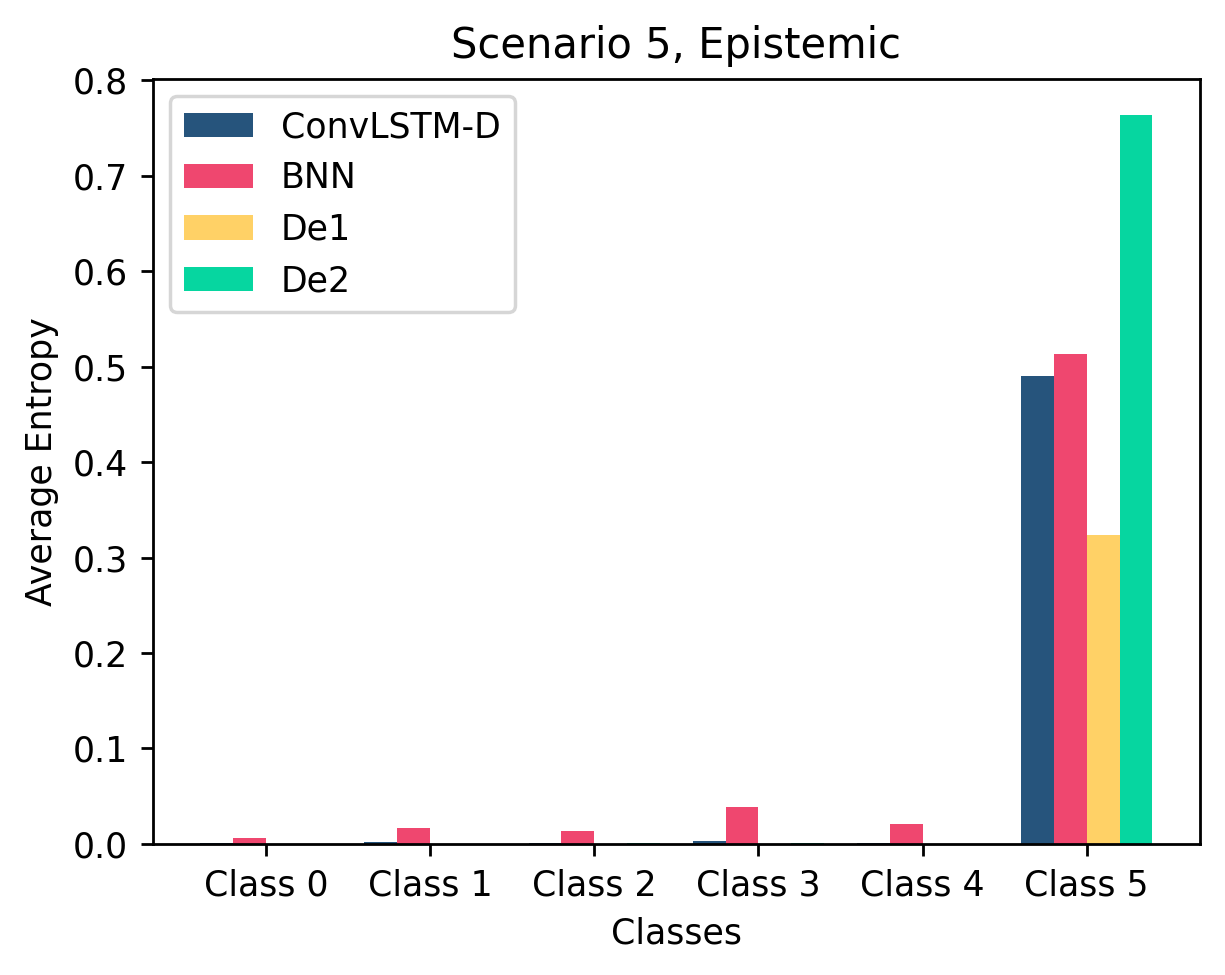}
		\caption{Scenario No.5}
	\end{subfigure}

	\caption{\textbf{Average entropy of the predictions for each class, computed on the examples in the validation set.}}
	\label{fig:16}
\end{figure}

As previously described, OOD data identification relies on the comparison between the entropy of the predictions and a given entropy threshold, which can be calculated in the validation phase.
For each epistemic uncertainty scenario, Figure~\ref{fig:16} reports the average entropy of the predictions for the examples in the validation set.
Each plot in the figure clearly shows the greater uncertainty of all models when predicting OOD data.
Indeed, the average entropy computed for the hold-out class in one scenario (e.g., class 0 in Scenario No.0) is consistently higher compared to the others.
BNN sets itself apart by providing slightly higher average entropy values for the non-hold-out classes.

The entropy of the predictions is then used to calculate the entropy threshold.
In our experiments, this is achieved by means of two alternative approaches: the "uncertainty outlier detection" method and the "uncertainty $f_1$ score" technique, giving rise to threshold $\tau_1$ and $\tau_2$, respectively.
These thresholds are displayed in Table~\ref{tab:4} and reveal an interesting and consistent pattern: $\tau_1$ is invariably lower than $\tau_2$ across all scenarios.
As a consequence, the adoption of $\tau_1$ at inference time is likely to result in a more cautious classification of the examples as unreliable.

\begin{table}[ht]
\centering
\setlength{\tabcolsep}{3pt} 
\renewcommand{\arraystretch}{1.5} 
\begin{tabular}{|c|cc|cc|cc|cc|}
\hline
\multirow{2}{*}{\textbf{Scenario}} & \multicolumn{2}{c|}{\textbf{ConvLSTM-D}} & \multicolumn{2}{c|}{\textbf{BNN}} & \multicolumn{2}{c|}{\textbf{De1}} & \multicolumn{2}{c|}{\textbf{De2}} \\ 
\cline{2-9}
 & $\boldsymbol{\tau_1}$ & $\boldsymbol{\tau_2}$ & $\boldsymbol{\tau_1}$ & $\boldsymbol{\tau_2}$ & $\boldsymbol{\tau_1}$ & $\boldsymbol{\tau_2}$ & $\boldsymbol{\tau_1}$ & $\boldsymbol{\tau_2}$ \\ 
\hline
No.0 & 0.0139 & 0.0205 & 0.0334 & 0.0400 & 0.0001 & 0.0017 & 0.0001 & 0.0016 \\
No.1 & 0.0150 & 0.0189 & 0.0902 & 0.5081 & 0.0153 & 0.0308 & 0.0098 & 0.0658 \\
No.2 & 0.0206 & 0.0209 & 0.0617 & 0.0849 & 0.0001 & 0.0034 & 0.0015 & 0.0033 \\
No.3 & 0.0088 & 0.0104 & 0.1453 & 0.1503 & 0.0006 & 0.0901 & 0.0016 & 0.0686 \\
No.4 & 0.0106 & 0.0130 & 0.1610 & 0.3539 & 0.0004 & 0.0070 & 0.0010 & 0.0883 \\
No.5 & 0.0036 & 0.0067 & 0.0287 & 0.3235 & 0.0001 & 0.0015 & 0.0005 & 0.0054 \\
\hline
\end{tabular}%
\caption{\textbf{Values of $\tau_1$ and $\tau_2$ computed for each model and epistemic scenario in the validation phase.}}
\label{tab:4}
\end{table}

\begin{table}[ht]
\centering
\setlength{\tabcolsep}{3pt}
\renewcommand{\arraystretch}{1.3}
\begin{tabular}{|l|c|c|c|c|c|c|c|c|}
\hline
\multirow{4}{*}{\textbf{Scenario}} & \multicolumn{8}{|c|}{\textbf{Model}} \\ 
\cline{2-9}
 & \multicolumn{2}{c|}{\textbf{ConvLSTM-D}} & \multicolumn{2}{|c|}{\textbf{BNN}} & \multicolumn{2}{|c|}{\textbf{De1}} & \multicolumn{2}{|c|}{\textbf{De2}} \\
\cline{2-9}

 & \begin{tabular}[c]{@{}c@{}} OOD \\ $\rightarrow$ UT$^\dagger$ \end{tabular} 
 & \begin{tabular}[c]{@{}c@{}} ID \\ $\rightarrow$ UT$^\ddagger$ \end{tabular} 
 & \begin{tabular}[c]{@{}c@{}} OOD \\ $\rightarrow$ UT \end{tabular} 
 & \begin{tabular}[c]{@{}c@{}} ID \\ $\rightarrow$ UT \end{tabular}
 & \begin{tabular}[c]{@{}c@{}} OOD \\ $\rightarrow$ UT \end{tabular} 
 & \begin{tabular}[c]{@{}c@{}} ID \\ $\rightarrow$ UT \end{tabular}
 & \begin{tabular}[c]{@{}c@{}} OOD \\ $\rightarrow$ UT \end{tabular} 
 & \begin{tabular}[c]{@{}c@{}} ID \\ $\rightarrow$ UT \end{tabular} \\
\hline
No.0 & 41.4\% & \textbf{5.3\%} & 41.0\% & 14.3\% & 95.9\% & 14.8\% & \textbf{96.9\%} & 13.5\% \\
No.1 & 81.4\% & 11.8\% & \textbf{98.0\%} & 16.9\% & 94.7\% & \textbf{9.0\%} & 95.8\% & 10.0\% \\
No.2 & 54.9\% & \textbf{2.9\%} & 47.0\% & 15.5\% & \textbf{98.7\%} & 15.7\% & 93.1\% & 4.5\% \\
No.3 & 80.1\% & 21.5\% & 65.4\% & \textbf{13.4\%} & 99.2\% & 18.3\% & \textbf{99.4\%} & 15.2\% \\
No.4 & 81.8\% & \textbf{8.9\%} & 74.6\% & 15.5\% & 77.2\% & 19.3\% & \textbf{100\%} & 12.4\% \\
No.5 & 88.6\% & \textbf{3.5\%} & 96.1\% & 13.9\% & 97.4\% & 10.5\% & \textbf{99.6\%} & 6.2\% \\
\hline
\textbf{Average} & 71.4\% & \textbf{9.0\%} & 70.4\% & 14.9\% & 93.9\% & 14.6\% & \textbf{97.5\%} & 10.3\% \\
\hline
\end{tabular}
\begin{tablenotes}
\footnotesize
\item $^\dagger$ OOD $\rightarrow$ UT: Percentage of out-of-distribution test examples labeled as untrustworthy.
\item $^\ddagger$ ID $\rightarrow$ UT: Percentage of in-distribution test examples labeled as untrustworthy.
\end{tablenotes}
\caption{\textbf{Percentage of OOD and ID data labeled as untrustworthy in the test phase by applying threshold $\tau_1$.}}
\label{tab:5}
\end{table}

\begin{table}[ht]
\centering
\setlength{\tabcolsep}{3pt}
\renewcommand{\arraystretch}{1.3}
\begin{tabular}{|l|c|c|c|c|c|c|c|c|}
\hline
\multirow{4}{*}{\textbf{Scenario}} & \multicolumn{8}{c|}{\textbf{Model}} \\ 
\cline{2-9}
 & \multicolumn{2}{c|}{\textbf{ConvLSTM-D}} & \multicolumn{2}{|c|}{\textbf{BNN}} & \multicolumn{2}{|c|}{\textbf{De1}} & \multicolumn{2}{|c|}{\textbf{De2}} \\
\cline{2-9}

  & \begin{tabular}[c]{@{}c@{}} OOD \\ $\rightarrow$ UT$^\dagger$ \end{tabular} 
 & \begin{tabular}[c]{@{}c@{}} ID \\ $\rightarrow$ UT$^\ddagger$ \end{tabular} 
 & \begin{tabular}[c]{@{}c@{}} OOD \\ $\rightarrow$ UT \end{tabular} 
 & \begin{tabular}[c]{@{}c@{}} ID \\ $\rightarrow$ UT \end{tabular}
 & \begin{tabular}[c]{@{}c@{}} OOD \\ $\rightarrow$ UT \end{tabular} 
 & \begin{tabular}[c]{@{}c@{}} ID \\ $\rightarrow$ UT \end{tabular}
 & \begin{tabular}[c]{@{}c@{}} OOD \\ $\rightarrow$ UT \end{tabular} 
 & \begin{tabular}[c]{@{}c@{}} ID \\ $\rightarrow$ UT \end{tabular} \\
\hline
No.0 & 36.2\% & 2.2\% & 39.5\% & 12.8\% & 74.0\% & \textbf{1.7\%} & \textbf{82.9\%} & 2.1\% \\
No.1 & 76.8\% & 9.3\% & 77.0\% & 2.6\% & 84.0\% & 4.6\% & \textbf{86.1\%} & \textbf{1.7\%} \\
No.2 & 54.7\% & 2.6\% & 41.6\% & 12.2\% & 75.7\% & 2.3\% & \textbf{83.8\%} & \textbf{1.1\%} \\
No.3 & 74.6\% & 20.9\% & 64.6\% & 13.1\% & 89.5\% & \textbf{3.4\%} & \textbf{90.3\%} & 3.5\% \\
No.4 & 76.5\% & 6.5\% & 61.3\% & 8.7\% & 65.6\% & 6.8\% & \textbf{96.2\%} & \textbf{2.1\%} \\
No.5 & 85.3\% & 0.4\% & 82.2\% & \textbf{0.3\%} & 92.4\% & 0.7\% & \textbf{97.1\%} & \textbf{0.3\%} \\
\hline
\textbf{Average} & 67.4\% & 7.0\% & 61.0\% & 8.3\% & 80.2\% & 3.3\% & \textbf{89.1\%} & \textbf{1.8\%} \\
\hline
\end{tabular}
\begin{tablenotes}
\footnotesize
\item $^\dagger$ OOD $\rightarrow$ UT: Percentage of out-of-distribution test examples labeled as untrustworthy.
\item $^\ddagger$ ID $\rightarrow$ UT: Percentage of in-distribution test examples labeled as untrustworthy.
\end{tablenotes}
\caption{\textbf{Percentage of OOD and ID data labeled as untrustworthy in the test phase by applying threshold $\tau_2$.}}
\label{tab:6}
\end{table}

By comparing the entropy threshold with the entropy of the predictions for each example in the test set, it is finally possible to evaluate the percentage of the out-of-distribution (OOD) data correctly labeled as untrustworthy, and in-distribution (ID) data erroneously flagged as such. 
The outcomes obtained for the competing models are presented in Tables~\ref{tab:5} and~\ref{tab:6} for $\tau_1$ and $\tau_2$, respectively.

From these results some empirical conclusions can be drawn.
First, the deep ensemble architectures surpass ConvLSTM-D and BNN in correctly identifying OOD data.
In particular, the De2 model demonstrates exceptional capability in detecting such instances, irrespective of the threshold applied.
In contrast, and with little exception, ConvLSTM-D and BNN exhibit notable limitations in addressing epistemic uncertainty, with their shortcomings particularly evident in Scenarios No.0 and No.2, where their performance lags behind De1 and De2 regardless of the threshold. 
With respect to the misclassification of ID data, the most favorable average outcome, under $\tau_1$, is achieved by ConvLSTM-D, followed by De2.
When applying $\tau_2$, instead, De2 and De1 emerge as the top-performing models.
Notice that, the quality of the predictions varies significantly across the scenarios.
These differences may reveal that the effectiveness of the untrustworthy analysis depends on the class designated as the hold-out class.
For example, in Scenario No.5 all models achieve notable results, identifying at least 88\% and 82\% of OOD data while misclassifying less than 15\% and 2\% of ID data, when $\tau_1$ and $\tau_2$ are, respectively, applied.
In contrast, in Scenarios No.0 and No.2 OOD data detection rates are considerably lower compared to Scenario No.5.
Despite these fluctuations, however, all models consistently identify a significant percentage of OOD data.

A noteworthy insight finally emerges from comparing the outcomes based on the adoption of $\tau_1$ and $\tau_2$.
When $\tau_1$ is employed (Table~\ref{tab:5}), all models achieve higher OOD data detection rates but misclassify a larger proportion of ID data, compared to $\tau_2$ (Table~\ref{tab:6}).
In particular, by applying $\tau_1$ the percentage of correctly classified OOD data increases of about 8\%, whereas for ID data it reduces of 7\%.
This finding aligns with expectations - as shown in Table~\ref{tab:4}, being consistently lower than $\tau_2$, $\tau_1$ is more conservative - and may guide the final choice of the threshold to employ.
On one hand, in critical applications prioritizing the identification of OOD data may justify the use of a stricter threshold, like $\tau_1$, as it ensures the correct detection of a larger amount of unreliable data. On the other hand, the greater misclassification of ID data may require the involvement of the human to recognize mislabeled ID data - a process that is both resource-intensive and costly.
Given the typical imbalance between ID and OOD data, also reflected in our scenarios (ID examples are five times OOD instances), even minor increases in ID data misclassification rate translate into a significant rise in post-detection human interventions.
This additional effort may be undesirable in terms of time and cost.
The adoption of $\tau_2$, first proposed in this study, appears the more pragmatic choice since, in practical applications, it minimizes the need for extensive manual re-evaluation.

\subsection{Aleatoric uncertainty scenarios}

To test the models in aleatoric uncertainty scenarios, four types of noise were injected into the data sets: Gaussian noise and three non-Gaussian types, i.e., Impulse, Rayleigh, and Weibull. 

\begin{figure}[H]
 \centering
 \includegraphics[width=0.9\linewidth]{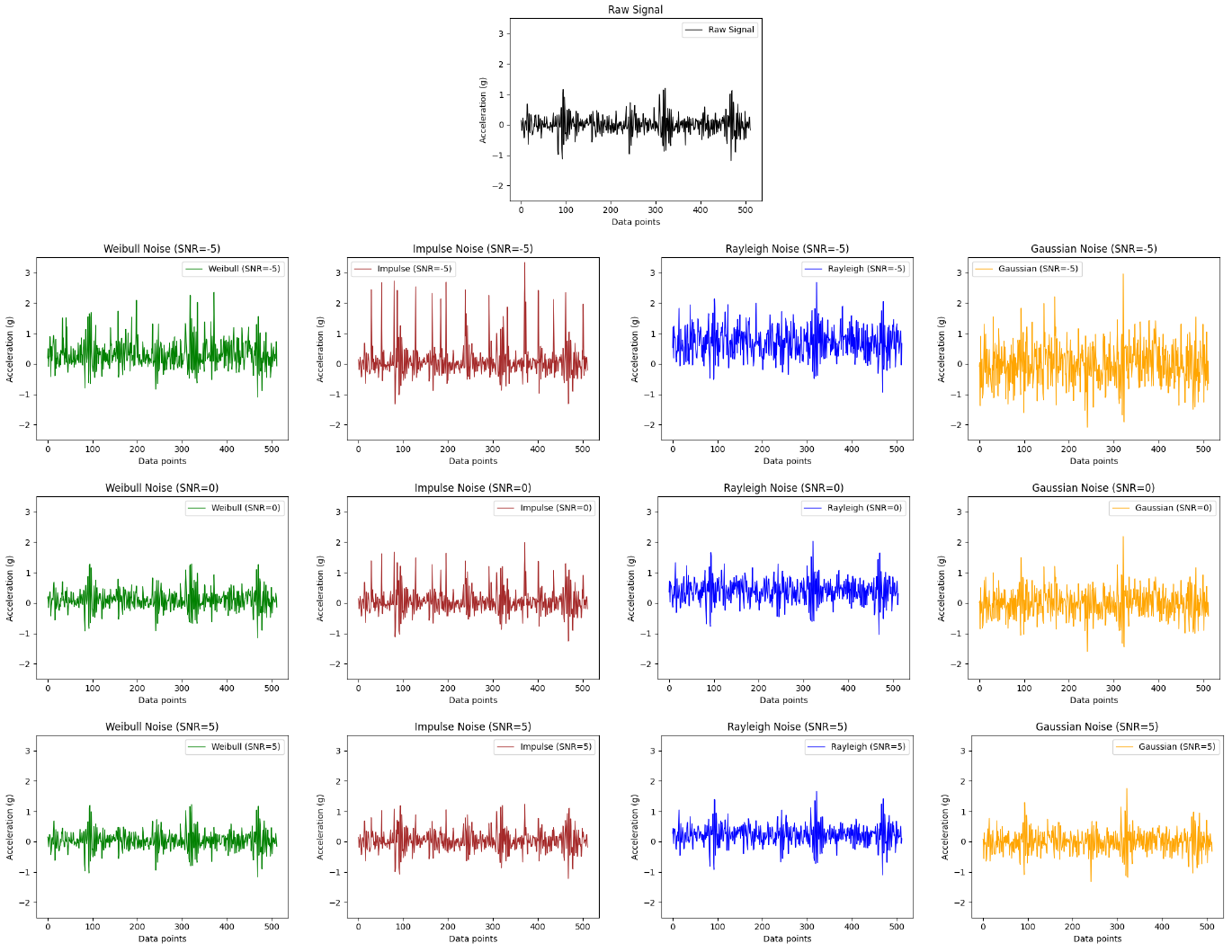}
 \caption{\textbf{Effect of noise types and intensities. The black signal is the original raw-signal. Green, brown, blue and yellow signals result from applying Weibull, Impulse, Rayleigh and Gaussian noise at different intensity levels (SNR = -5 dB, 0 dB, 5 dB).}}
 \label{fig:17}
\end{figure}

Noise was applied at varying strength levels corresponding to signal-to-noise ratio (SNR) values of -5 dB, 0 dB, and 5 dB, to represent strong, medium, and slight noise, respectively.
This process yielded 12 aleatoric uncertainty scenarios derived from the combination of the four noise types and the three intensity levels. 
Noise was added to 20\% of the data, randomly drawn from both the validation and the test set.
To maintain consistency, a fixed random seed was employed to guarantee the use of the same noise-injected data in all scenarios.
To visualize how the noise type and its intensity affected the data, Figure~\ref{fig:17} shows the perturbation induced by the addition of noise on one original, non-noisy, example (raw signal).

As for epistemic uncertainty, $\tau_1$ and $\tau_2$ were determined on the validation examples, for all noisy scenarios.
These values are detailed in Table~\ref{tab:7}.
Notably, $\tau_1$ remains constant across varying noise levels since it is computed using outlier detection on ID (non-noisy) data, which is uniform across the scenarios. 
In contrast, $\tau_2$ fluctuates with the noise intensity, as it is calculated on both ID and OOD data from the validation set. 
Specifically, greater noise leads to increased OOD prediction entropy, thereby elevating $\tau_2$. 
Furthermore, $\tau_1$ consistently stays below $\tau_2$, particularly at high noise levels, mirroring the conservative nature of $\tau_1$ observed in epistemic uncertainty settings.

\begin{table}[ht]
\setlength{\tabcolsep}{3pt}
\renewcommand{\arraystretch}{1.2}
\centering
\begin{tabular}{|l|l|clllllll|}
\hline
\multicolumn{1}{|c|}{\multirow{3}{*}{\begin{tabular}[c]{@{}c@{}} \textbf{Noise}\\ \textbf{type} \end{tabular}}} & \multicolumn{1}{c|}{\multirow{3}{*}{\begin{tabular}[c]{@{}c@{}} \textbf{Noise}\\ \textbf{level} \end{tabular}}} & \multicolumn{8}{c|}{\textbf{Model}}  \\ \cline{3-10} 
\multicolumn{1}{|c|}{} & \multicolumn{1}{c|}{}  & \multicolumn{2}{c|}{\textbf{ConvLSTM-D}}  & \multicolumn{2}{c|}{\textbf{BNN}} & \multicolumn{2}{c|}{\textbf{De1}} & \multicolumn{2}{c|}{\textbf{De1}} \\ \cline{3-10} 
\multicolumn{1}{|c|}{} & \multicolumn{1}{c|}{}  & \multicolumn{1}{c|}{$\tau_1$}  & \multicolumn{1}{c|}{$\tau_2$} & \multicolumn{1}{c|}{$\tau_1$}  & \multicolumn{1}{c|}{$\tau_2$} & \multicolumn{1}{c|}{$\tau_1$}  & \multicolumn{1}{c|}{$\tau_2$} & \multicolumn{1}{c|}{$\tau_1$}  & $\tau_2$ \\ \hline

\multirow{3}{*}{Weilbul}  & Strong  & \multicolumn{1}{c|}{\multirow{12}{*}{0.0236}} & \multicolumn{1}{l|}{0.0791} & \multicolumn{1}{l|}{\multirow{12}{*}{0.2965}} & \multicolumn{1}{l|}{0.3485} & \multicolumn{1}{l|}{\multirow{12}{*}{0.0002}} & \multicolumn{1}{l|}{0.1488} & \multicolumn{1}{l|}{\multirow{12}{*}{0.0024}} & \multicolumn{1}{l|}{0.3452} \\
 & Medium  & \multicolumn{1}{c|}{} & \multicolumn{1}{l|}{0.0284} & \multicolumn{1}{l|}{} & \multicolumn{1}{l|}{0.2876} & \multicolumn{1}{l|}{} & \multicolumn{1}{l|}{0.0248} & \multicolumn{1}{l|}{} & \multicolumn{1}{l|}{0.2166} \\
 & Slight  & \multicolumn{1}{c|}{} & \multicolumn{1}{l|}{0.0151} & \multicolumn{1}{l|}{} & \multicolumn{1}{l|}{0.1046} & \multicolumn{1}{l|}{} & \multicolumn{1}{l|}{0.0367} & \multicolumn{1}{l|}{} & \multicolumn{1}{l|}{0.1481}\\ \cline{1-2} \cline{4-4} \cline{6-6} \cline{8-8} \cline{10-10} 

 \multirow{3}{*}{Impulse}  & Strong & \multicolumn{1}{c|}{} & \multicolumn{1}{l|}{0.0745} & \multicolumn{1}{l|}{} & \multicolumn{1}{l|}{0.3347} & \multicolumn{1}{l|}{} & \multicolumn{1}{l|}{0.0381} & \multicolumn{1}{l|}{} & \multicolumn{1}{l|}{0.1853}\\
 & Medium & \multicolumn{1}{c|}{} & \multicolumn{1}{l|}{0.0489} & \multicolumn{1}{l|}{} & \multicolumn{1}{l|}{0.2063} & \multicolumn{1}{l|}{} & \multicolumn{1}{l|}{0.0270} & \multicolumn{1}{l|}{} & \multicolumn{1}{l|}{0.0071}\\
 & Slight & \multicolumn{1}{c|}{} & \multicolumn{1}{l|}{0.0202} & \multicolumn{1}{l|}{} & \multicolumn{1}{l|}{0.0035} & \multicolumn{1}{l|}{} & \multicolumn{1}{l|}{0.0004} & \multicolumn{1}{l|}{} & \multicolumn{1}{l|}{0.0083}\\ \cline{1-2} \cline{4-4} \cline{6-6} \cline{8-8} \cline{10-10} 

 \multirow{3}{*}{Rayleigh} & Strong & \multicolumn{1}{c|}{} & \multicolumn{1}{l|}{0.0792} & \multicolumn{1}{l|}{} & \multicolumn{1}{l|}{0.3952} & \multicolumn{1}{l|}{} & \multicolumn{1}{l|}{0.4731} & \multicolumn{1}{l|}{} & \multicolumn{1}{l|}{0.2531}\\
 & Medium & \multicolumn{1}{c|}{} & \multicolumn{1}{l|}{0.0761} & \multicolumn{1}{l|}{} & \multicolumn{1}{l|}{0.4640} & \multicolumn{1}{l|}{} & \multicolumn{1}{l|}{0.5559} & \multicolumn{1}{l|}{} & \multicolumn{1}{l|}{0.2077}\\
 & Slight & \multicolumn{1}{c|}{} & \multicolumn{1}{l|}{0.0219} & \multicolumn{1}{l|}{} & \multicolumn{1}{l|}{0.3007} & \multicolumn{1}{l|}{} & \multicolumn{1}{l|}{0.0429} & \multicolumn{1}{l|}{} & \multicolumn{1}{l|}{0.1187}\\ \cline{1-2} \cline{4-4} \cline{6-6} \cline{8-8} \cline{10-10} 

 \multirow{3}{*}{Gaussian} & Strong & \multicolumn{1}{c|}{} & \multicolumn{1}{l|}{0.0573} & \multicolumn{1}{l|}{} & \multicolumn{1}{l|}{0.3571} & \multicolumn{1}{l|}{} & \multicolumn{1}{l|}{0.0800} & \multicolumn{1}{l|}{} & \multicolumn{1}{l|}{0.4394}\\
 & Medium & \multicolumn{1}{c|}{} & \multicolumn{1}{l|}{0.0365} & \multicolumn{1}{l|}{} & \multicolumn{1}{l|}{0.3485} & \multicolumn{1}{l|}{} & \multicolumn{1}{l|}{0.0017} & \multicolumn{1}{l|}{} & \multicolumn{1}{l|}{0.0844}\\
 & Slight & \multicolumn{1}{c|}{} & \multicolumn{1}{l|}{0.0117} & \multicolumn{1}{l|}{} & \multicolumn{1}{l|}{0.2516} & \multicolumn{1}{l|}{} & \multicolumn{1}{l|}{0.0007} & \multicolumn{1}{l|}{} & \multicolumn{1}{l|}{0.0021} \\ \hline
\end{tabular}%
\caption{\textbf{Values of $\tau_1$ and $\tau_2$ computed for each model and aleatoric scenario in the validation phase.}}
\label{tab:7}
\end{table}

\begin{figure}[ht]
 \centering
 \includegraphics[width=\linewidth]{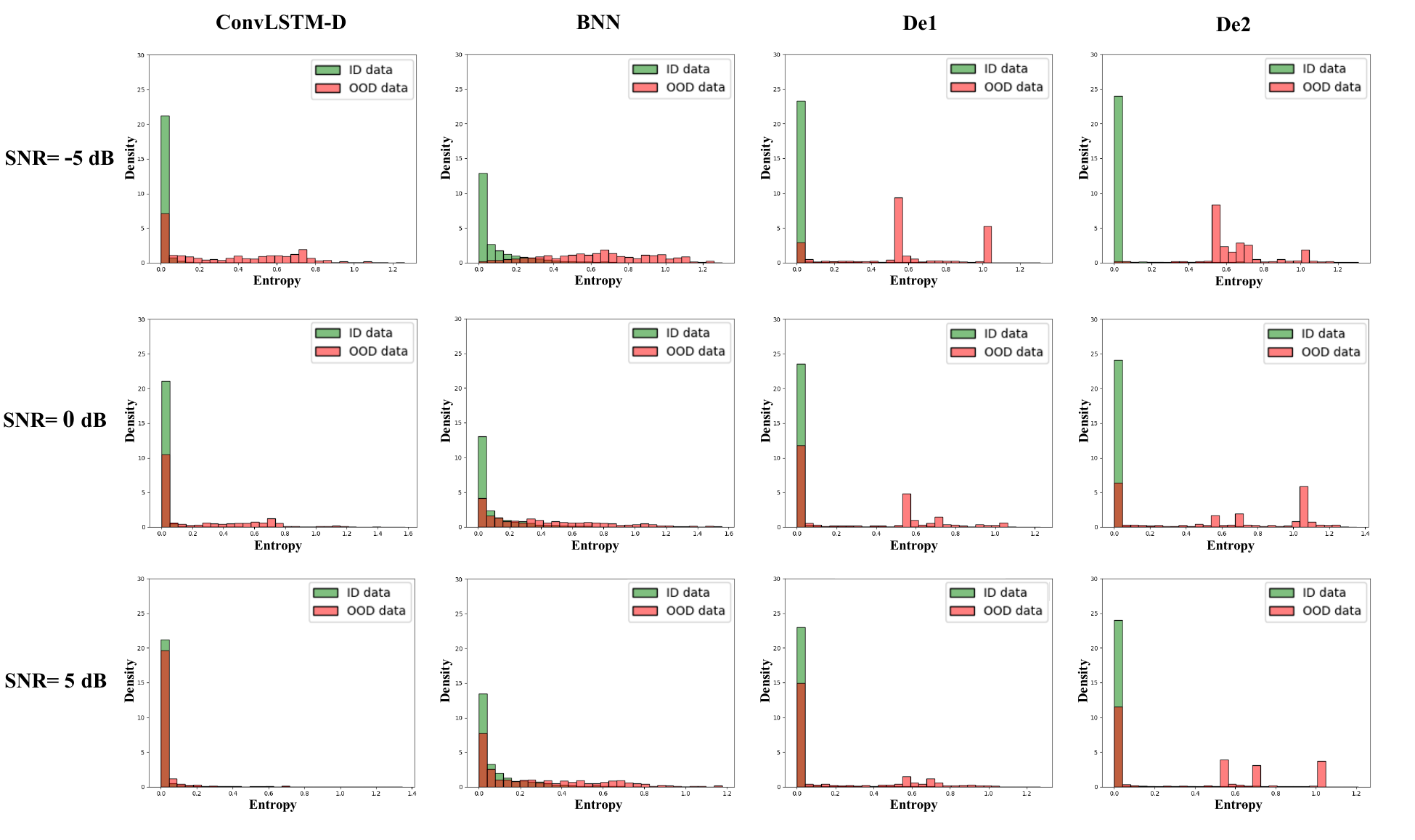}
 \caption{\textbf{Entropy distribution for ID (green) and OOD (red) data when Weibull noise is applied, for different noise levels (SNR = -5 dB, 0 dB, and 5 dB) across the four models (ConvLSTM-D, BNN, De1, and De2).}}
 \label{fig:18}
\end{figure}

To illustrate how the distribution of entropy for ID and OOD data varies under different noise levels, Figure~\ref{fig:18} presents this information on the test set for the Weibull noise, across all models.
The figure demonstrates that, as noise intensity increases (from SNR = 5 dB to SNR = -5 dB), the entropy distribution of OOD data consistently shifts rightward, toward lower entropy values.
This aligns with the expectation that detecting OOD data under aleatoric uncertainty conditions becomes progressively easier with higher noise levels.
Conversely, at lower noise levels the entropy distributions of ID and OOD data are closely aligned. 
A similar trend is observed for Impulse, Rayleigh, and Gaussian noise types, as shown in Figures~\ref{fig:B1}, \ref{fig:B2}, and~\ref{fig:B3} in Appendix~\ref{app:B}.

Similarly to the epistemic uncertainty analysis, the effectiveness of flagging OOD data as untrustworthy and the proportion of ID data erroneously classified were evaluated.
This was achieved by comparing the prediction entropy on the test examples against the thresholds established in the validation phase.
The percentages of OOD and ID data labeled as untrustworthy were calculated across all the scenarios and the results are presented in Tables~\ref{tab:8} and~\ref{tab:9} for $\tau_1$ and $\tau_2$, respectively.
The tables reveal that OOD detection improves significantly under strong noise conditions, regardless of the threshold used or the type of noise.
Conversely, at low noise levels the uncertainty analysis performance declines sharply.
This is attributable to the greater similarity between ID and OOD data under slight noise conditions, leading to comparable entropy values and hampering OOD detection.
By contrast, strong noise amplifies the prediction entropy of OOD data relative to ID data, facilitating more accurate detection. This explains the notable difference in OOD detection performance between high noise (SNR = -5 dB) and low noise (SNR = 5 dB) scenarios. A similar, albeit less pronounced, pattern is observed when comparing strong to medium noise and medium to slight noise settings.

\begin{table}[ht]
\setlength{\tabcolsep}{3pt}
\renewcommand{\arraystretch}{1.2}
\centering
\begin{tabular}{|l|l|llllllll|}
\hline
\multicolumn{1}{|c|}{\multirow{4}{*}{\begin{tabular}[c]{@{}c@{}} \textbf{Noise}\\ \textbf{type} \end{tabular}}} & \multicolumn{1}{c|}{\multirow{4}{*}{\begin{tabular}[c]{@{}c@{}} \textbf{Noise}\\ \textbf{level} \end{tabular}}} & \multicolumn{8}{c|}{\textbf{Model}}  \\ \cline{3-10} 
\multicolumn{1}{|c|}{} & \multicolumn{1}{c|}{}  & \multicolumn{2}{c|}{\textbf{ConvLSTM-D}}  & \multicolumn{2}{c|}{\textbf{BNN}} & \multicolumn{2}{c|}{\textbf{De1}} & \multicolumn{2}{c|}{\textbf{De1}} \\ \cline{3-10} 
\multicolumn{1}{|c|}{} & \multicolumn{1}{c|}{}  
& \multicolumn{1}{c|}{\begin{tabular}[c]{@{}c@{}} OOD \\ $\rightarrow$ UT$^\dagger$ \end{tabular}}  
& \multicolumn{1}{c|}{\begin{tabular}[c]{@{}c@{}} ID \\ $\rightarrow$ UT$^\ddagger$ \end{tabular}} 
& \multicolumn{1}{c|}{\begin{tabular}[c]{@{}c@{}} OOD \\ $\rightarrow$ UT \end{tabular}}
& \multicolumn{1}{c|}{\begin{tabular}[c]{@{}c@{}} ID \\ $\rightarrow$ UT \end{tabular}} 
& \multicolumn{1}{c|}{\begin{tabular}[c]{@{}c@{}} OOD \\ $\rightarrow$ UT \end{tabular}}  
& \multicolumn{1}{c|}{\begin{tabular}[c]{@{}c@{}} ID \\ $\rightarrow$ UT \end{tabular}} 
& \multicolumn{1}{c|}{\begin{tabular}[c]{@{}c@{}} OOD \\ $\rightarrow$ UT \end{tabular}}  
& \begin{tabular}[c]{@{}c@{}} ID \\ $\rightarrow$ UT \end{tabular} \\ \hline

\multirow{3}{*}{Weilbul}  & Strong & \multicolumn{1}{c|}{72.5\%}  & \multicolumn{1}{l|}{\multirow{12}{*}{10.1\%}} & \multicolumn{1}{l|}{82.8\%} & \multicolumn{1}{l|}{\multirow{12}{*}{10.2\%}} & \multicolumn{1}{l|}{96.7\%} & \multicolumn{1}{l|}{\multirow{12}{*}{19.1\%}} & \multicolumn{1}{l|}{\textbf{100\%}} & \multicolumn{1}{l|}{\multirow{12}{*}{\textbf{9.8\%}}} \\\
 & Medium  & \multicolumn{1}{c|}{38.9\%} & \multicolumn{1}{l|}{} & \multicolumn{1}{l|}{48.5\%} & \multicolumn{1}{l|}{} & \multicolumn{1}{l|}{70.4\%} & \multicolumn{1}{l|}{} & \multicolumn{1}{l|}{\textbf{74.3\%}} & \multicolumn{1}{l|}{} \\
 & Slight  & \multicolumn{1}{c|}{19.9\%} & \multicolumn{1}{l|}{} & \multicolumn{1}{l|}{22.5\%} & \multicolumn{1}{l|}{} & \multicolumn{1}{l|}{45.6\%} & \multicolumn{1}{l|}{} & \multicolumn{1}{l|}{\textbf{59.0\%}} & \multicolumn{1}{l|}{} \\ \cline{1-3} \cline{5-5} \cline{7-7} \cline{9-9} 

 \multirow{3}{*}{Impulse}  & Strong & \multicolumn{1}{c|}{74.0\%} & \multicolumn{1}{l|}{} & \multicolumn{1}{l|}{75.0\%} & \multicolumn{1}{l|}{} & \multicolumn{1}{l|}{84.8\%} & \multicolumn{1}{l|}{} & \multicolumn{1}{l|}{\textbf{87.4\%}} & \multicolumn{1}{l|}{} \\
 & Medium & \multicolumn{1}{c|}{49.5\%} & \multicolumn{1}{l|}{} & \multicolumn{1}{l|}{44.1\%} & \multicolumn{1}{l|}{} & \multicolumn{1}{l|}{\textbf{81.2\%}} & \multicolumn{1}{l|}{} & \multicolumn{1}{l|}{75.2\%} & \multicolumn{1}{l|}{}\\
 & Slight & \multicolumn{1}{c|}{28.6\%} & \multicolumn{1}{l|}{} & \multicolumn{1}{l|}{18.1\%} & \multicolumn{1}{l|}{} & \multicolumn{1}{l|}{\textbf{69.0\%}} & \multicolumn{1}{l|}{} & \multicolumn{1}{l|}{47.6\%} & \multicolumn{1}{l|}{}\\ \cline{1-3} \cline{5-5} \cline{7-7} \cline{9-9} 

 \multirow{3}{*}{Rayleigh} & Strong & \multicolumn{1}{c|}{73.6\%} & \multicolumn{1}{l|}{} & \multicolumn{1}{l|}{85.0\%} & \multicolumn{1}{l|}{} & \multicolumn{1}{l|}{94.5\%} & \multicolumn{1}{l|}{} & \multicolumn{1}{l|}{\textbf{100\%}} & \multicolumn{1}{l|}{}\\
 & Medium & \multicolumn{1}{c|}{51.8\%} & \multicolumn{1}{l|}{} & \multicolumn{1}{l|}{58.8\%} & \multicolumn{1}{l|}{} & \multicolumn{1}{l|}{78.0\%} & \multicolumn{1}{l|}{} & \multicolumn{1}{l|}{\textbf{87.1\%}} & \multicolumn{1}{l|}{}\\
 & Slight & \multicolumn{1}{c|}{38.0\%} & \multicolumn{1}{l|}{} & \multicolumn{1}{l|}{27.4\%} & \multicolumn{1}{l|}{} & \multicolumn{1}{l|}{44.3\%} & \multicolumn{1}{l|}{} & \multicolumn{1}{l|}{\textbf{58.9\%}} & \multicolumn{1}{l|}{}\\ \cline{1-3} \cline{5-5} \cline{7-7} \cline{9-9} 

 \multirow{3}{*}{Gaussian} & Strong & \multicolumn{1}{c|}{74.5\%} & \multicolumn{1}{l|}{} & \multicolumn{1}{l|}{73.9\%} & \multicolumn{1}{l|}{} & \multicolumn{1}{l|}{78.9\%} & \multicolumn{1}{l|}{} & \multicolumn{1}{l|}{\textbf{90.1\%}} & \multicolumn{1}{l|}{}\\
 & Medium & \multicolumn{1}{c|}{48.5\%} & \multicolumn{1}{l|}{} & \multicolumn{1}{l|}{47.4\%} & \multicolumn{1}{l|}{} & \multicolumn{1}{l|}{72.9\%} & \multicolumn{1}{l|}{} & \multicolumn{1}{l|}{\textbf{83.7\%}} & \multicolumn{1}{l|}{}\\
 & Slight & \multicolumn{1}{c|}{18.5\%} & \multicolumn{1}{l|}{} & \multicolumn{1}{l|}{32.5\%} & \multicolumn{1}{l|}{} & \multicolumn{1}{l|}{\textbf{66.1\%}} & \multicolumn{1}{l|}{} & \multicolumn{1}{l|}{52.1\%} & \multicolumn{1}{l|}{}\\ \hline
\end{tabular}%
\begin{tablenotes}
\footnotesize
\item $^\dagger$ OOD $\rightarrow$ UT: Percentage of out-of-distribution test examples labeled as untrustworthy.
\item $^\ddagger$ ID $\rightarrow$ UT: Percentage of in-distribution test examples labeled as untrustworthy.
\end{tablenotes}
\caption{\textbf{Percentage of OOD and ID data labeled as untrustworthy in the test phase by applying threshold $\tau_1$.}}
\label{tab:8}
\end{table}

\begin{table}[ht]
\setlength{\tabcolsep}{3pt}
\renewcommand{\arraystretch}{1.3}
\centering
\begin{tabular}{|l|l|c|c|c|c|c|c|c|c|}
\hline
\multicolumn{1}{|c|}{\multirow{4}{*}{\begin{tabular}[c]{@{}c@{}} \textbf{Noise}\\ \textbf{type} \end{tabular}}} & \multicolumn{1}{c|}{\multirow{4}{*}{\begin{tabular}[c]{@{}c@{}} \textbf{Noise}\\ \textbf{level} \end{tabular}}} &
\multicolumn{8}{c|}{\textbf{Model}} \\
\cline{3-10}
 & & \multicolumn{2}{c|}{\textbf{ConvLSTM-D}} & \multicolumn{2}{|c|}{\textbf{BNN}} & \multicolumn{2}{|c|}{\textbf{De1}} & \multicolumn{2}{|c|}{\textbf{De2}} \\
\cline{3-10}
 & & \begin{tabular}[c]{@{}c@{}} OOD \\ $\rightarrow$ UT$^\dagger$ \end{tabular} 
 & \begin{tabular}[c]{@{}c@{}} ID \\ $\rightarrow$ UT$^\ddagger$ \end{tabular} 
 & \begin{tabular}[c]{@{}c@{}} OOD \\ $\rightarrow$ UT \end{tabular} 
 & \begin{tabular}[c]{@{}c@{}} ID \\ $\rightarrow$ UT \end{tabular}
 & \begin{tabular}[c]{@{}c@{}} OOD \\ $\rightarrow$ UT \end{tabular} 
 & \begin{tabular}[c]{@{}c@{}} ID \\ $\rightarrow$ UT \end{tabular}
 & \begin{tabular}[c]{@{}c@{}} OOD \\ $\rightarrow$ UT \end{tabular} 
 & \begin{tabular}[c]{@{}c@{}} ID \\ $\rightarrow$ UT \end{tabular} \\
\hline
\multirow{3}{*}{Weibull} & Strong & 66.3\% & 3.0\% & 62.2\% & 6.9\% & 84.5\% & 2.0\% & \textbf{97.6\%} & \textbf{1.4\%} \\
 & Medium & 36.6\% & 7.7\% & 44.8\% & 10.1\% & 62.5\% & 4.1\% & \textbf{67.1\%} & \textbf{1.7\%} \\
 & Slight & 32.8\% & 20.0\% & 42.4\% & 29.4\% & 37.1\% & 2.9\% & \textbf{51.4\%} & \textbf{2.4\%} \\
\hline
\multirow{3}{*}{Impulse} & Strong & 66.4\% & 2.8\% & 70.8\% & 7.7\% & 76.0\% & 3.7\% & \textbf{79.9\%} & \textbf{1.5\%} \\
 & Medium & 45.7\% & \textbf{3.8\%} & 49.2\% & 17.3\% & 70.7\% & 4.0\% & \textbf{71.7\%} & 4.6\% \\
 & Slight & 30.1\% & 11.1\% & 35.3\% & 25.6\% & \textbf{62.2\%} & 15.2\% & 40.1\% & \textbf{4.6\%} \\
\hline
\multirow{3}{*}{Rayleigh} & Strong & 63.1\% & 3.2\% & 77.1\% & 5.2\% & 86.3\% & \textbf{1.1\%} & \textbf{98.3\%} & 1.6\% \\
 & Medium & 44.4\% & 3.4\% & 52.0\% & 2.9\% & 76.0\% & \textbf{0.7\%} & \textbf{83.6\%} & 7.2\% \\
 & Slight & 37.4\% & 9.3\% & 32.2\% & 9.1\% & 36.9\% & \textbf{2.7\%} & \textbf{51.5\%} & \textbf{2.7\%} \\
\hline
\multirow{3}{*}{Gaussian} & Strong & 67.6\% & 3.0\% & 68.5\% & 5.8\% & 72.5\% & 2.4\% & \textbf{82.1\%} & \textbf{1.4\%} \\
 & Medium & 45.0\% & 4.3\% & 42.1\% & 6.0\% & 68.2\% & 8.5\% & \textbf{78.4\%} & \textbf{3.0\%} \\
 & Slight & 37.8\% & 26.0\% & 33.4\% & 13.1\% & 46.2\% & \textbf{4.4\%} & \textbf{51.8\%} & 11.7\% \\
\hline
\end{tabular}
\begin{tablenotes}
\footnotesize
\item $^\dagger$ OOD $\rightarrow$ UT: Percentage of out-of-distribution test examples labeled as untrustworthy.
\item $^\ddagger$ ID $\rightarrow$ UT: Percentage of in-distribution test examples labeled as untrustworthy.
\end{tablenotes}
\caption{\textbf{Percentage of OOD and ID data labeled as untrustworthy in the test phase by applying threshold $\tau_2$.}}
\label{tab:9}
\end{table}

While noise intensity affects OOD detection across all architectures, the extent of this impact varies. Under strong noise conditions, and for both thresholds, De1 and De2 demonstrate exceptional performance, identifying 80\% to 100\% of OOD data in most scenarios, whereas ConvLSTM-D and BNN achieve a lower detection range of 60\% to 80\%. At medium noise levels, De1 and De2 maintain their superiority, detecting over 60\% of OOD data, compared to less than 50\% detected by ConvLSTM-D and BNN across all noise types. Even in slight noise conditions, deep ensemble-based architectures continue to detect a significant portion of OOD data.

Considering noise type, OOD data influenced by Weibull and Rayleigh noise are more detectable by De1 and De2 under strong noise, regardless of the threshold applied, whereas ConvLSTM-D remains unaffected by variations in noise type in such conditions. The interplay of noise type and intensity collectively shapes the performance of the architectures, as reflected in the results of De1 for both $\tau_1$ and $\tau_2$. Under strong noise, De1 excels with Weibull and Rayleigh noise but shows greater sensitivity to these types as noise diminishes, with Impulse and Gaussian noise yielding superior performance in slight noise scenarios.
The percentage of ID data incorrectly flagged as untrustworthy varies notably depending on whether $\tau_1$ or $\tau_2$ is applied. Table \ref{tab:8}, based on $\tau_1$, reveals that De1 has the highest error rate (19\%), while other models hover around 10\%.
In contrast, under $\tau_2$, particularly in slight noise conditions where ID and OOD data are more alike, the error rate increases significantly. As shown in Table \ref{tab:9}, ConvLSTM-D and BNN mistakenly classify over 20\% of ID data as untrustworthy in these conditions, whereas De1 and De2 demonstrate comparatively better performance.

Another interesting finding emerges when comparing Tables \ref{tab:8} and \ref{tab:9}. In strong and medium noise scenarios, $\tau_1$ tends to flag slightly more OOD data as untrustworthy but also results in a higher rate of false positives, misclassifying ID data. This behavior well reflects the trade-off inherent of a conservative threshold, as noted in the former discussion of epistemic uncertainty. In particular, while conservative thresholds enhance the detection of OOD data and can be, therefore, advantageous for critical applications, they also increase the likelihood of erroneously labeling ID data as unreliable. This, in turn, might require the involvement of experts to distinguish genuine OOD from misclassified ID data, leading to increased time and effort.

Overall, it is evident that De2 generally outperforms other architectures, particularly under strong noise conditions and across all noise types. Indeed, it achieves higher rates of OOD detection while producing lower false positive rates.
De1 performs similarly well in identifying OOD data, whereas ConvLSTM-D and BNN exhibit comparatively weaker performance in this regard.
This finding deserves attention since it sets both De1 and De2, which demonstrated notable OOD detection capabilities also in epistemic uncertainty scenarios, as the most robust candidates among the proposed architectures.

\subsection{Computational time}

One of the challenges of uncertainty-aware DL models is their computational cost.
All experiments were conducted using Google Colab, which provided the required computational resources. Specifically, the platform used an Intel(R) Xeon(R) CPU @ 2.20GHz, 2 processors, with 1 core each, and 56 MB L3 cache. The system had a total of 12.67 GB of RAM, with approximately 9.75 GB available for model training and testing. All architectures were trained for 25 epochs.

To evaluate the computational effort, for each model both the training and the prediction time were measured. These values are listed in Table \ref{tab:10}, where the prediction time refers to the time required to generate the vectors of predictions for the entire test set. 
As can be noted, the training for De1 and De2 is significantly faster compared to ConvLSTM-D and BNN. 
This is due to the fact that for the proposed deep ensemble architectures four base learners must be trained.
On the contrary,  ConvLSTM-D and BNN are approximately 3 times slower in the prediction phase since, for these models, predictions need to be repeated $K$ (in our tests $K=10$) times. 
It is worthwhile to mention that, in practical applications the inference time plays a crucial role since faster prediction implies faster detection of faults and, consequently, more timely interventions. 
Based on this, despite their longer training time, De1 and De2 are therefore preferable. 
This evidence, combined with the previous results which demonstrate the superiority of the deep ensemble models for OOD detection in both epistemic and aleatoric uncertainty scenarios, makes De1 and De2 the most effective and efficient architectures, as well as the most promising candidates for future investigation and development.

\begin{table}[ht]
\centering
\setlength{\tabcolsep}{3pt}
\renewcommand{\arraystretch}{1.3}
\begin{tabular}{|l|r|r|r|r|}
\hline
\textbf{Time (s)} & \textbf{ConvLSTM-D} & \textbf{BNN} & \textbf{De1} & \textbf{De2} \\
\hline
Training   & \textbf{239.55} & 266.18 & 688.44 & 639.69 \\
Prediction & 13.73  & 15.27  & 5.66   & \textbf{5.50} \\
\hline
\end{tabular}
\caption{\textbf{Computational time (seconds) for training and prediction.}}
\label{tab:10}
\end{table}

\section{Conclusions and future works}
\label{sec:conclusions}

In this research, we proposed an extensive comparative study of different uncertainty-aware DL architectures for OOD detection in fault diagnosis, using the CWRU dataset. 
These architectures are Convolutional LSTM with sampling by dropout (ConvLSTM-D), Bayesian neural network (BNN), and deep ensemble. 
For the latter, two configurations of base learners were examined: one employing identical base learners (De1) and the other utilizing distinct base learners (De2).
Both epistemic and aleatoric uncertainty scenarios were investigated.
Regarding the latter, unlike previous studies, non-Gaussian noises (Weibull, Impulse, and Rayleigh) were considered, as well as Gaussian noise.
Furthermore, two entropy thresholds were employed for the identification of OOD data: one ($\tau_1$) based on the uncertainty outlier detection, and the other ($\tau_2$), here proposed for the first time, based on the uncertainty $f_1$ score.

In epistemic uncertainty scenarios we found that, on average, all DL models successfully detected the majority of OOD data regardless of the chosen entropy threshold. However, deep ensemble models, especially De2, outperformed ConvLSTM-D and BNN in OOD detection across both threshold choices.
With respect to the misclassification of ID data, ConvLSTM-D and De2 achieved superior results when $\tau_1$ was applied, whereas De2 and De1 performed better with $\tau_2$.
The results from the epistemic uncertainty scenarios revealed that, on average, $\tau_1$ flagged more OOD data as untrustworthy than $\tau_2$, albeit at the cost of mistakenly classifying more ID data as untrustworthy.

For aleatoric uncertainty scenarios, the noise level was shown to impact the performance of OOD detection. 
As expected, a lower SNR made the entropy values computed for OOD and ID data closer. 
This hindered the models' ability to effectively distinguish between OOD and ID data.
Nevertheless, De2 and De1 exhibited a smaller performance decline on average, making them better candidates for OOD detection, regardless of the threshold choice.
Based on our experiments, both noise type and noise level jointly influenced OOD detection performance, with variations in the noise type determining the extent to which noise level impacts the models' performance. 
Consistently with epistemic uncertainty, a similar conservative behavior was observed when selecting  $\tau_1$ instead of $\tau_2$. 
Finally, in terms of computational cost, De1 and De2 were faster at prediction but slower in the training phase. 

Based on our comparative study some implications can be derived and used as practical guidelines for real-world fault diagnosis applications in rotating machinery.
First, the combination of faster predictions and superior performance under both types of uncertainty makes the deep ensemble models, especially De2, more suitable for real-world use cases.
Moreover, as shown by our experiments, the types and levels of noise considerably impact the results.
Therefore, when applying the presented analysis to real-world scenarios, it would be recommended to anticipate and address noise during data collection, to improve the reliability of the model in the presence of aleatoric uncertainty.
Furthermore, the selection of the entropy threshold should correspond to the criticality of the application.
For highly critical domains where robust OOD detection is crucial, practitioners are encouraged to adopt a conservative threshold ($\tau_1$).
Conversely, for less critical tasks, a less conservative threshold ($\tau_2$) is recommended, as it helps to reduce false positives and streamline post-processing efforts.
By leveraging these considerations, researchers and practitioners can successfully adapt and build upon our findings for their future work on fault diagnosis applications.

It is worthwhile to notice that, the comparison among the different DL models was performed on a unique set of data, i.e. the CWRU dataset.
This choice was motivated by two main reasons.
First, the CWRU dataset is a widely recognized benchmark in fault diagnosis research.
Second, the variety of fault classes represented therein allowed for the generation of different epistemic uncertainty use cases. 
By resorting to this dataset, therefore, we have been able to address our goal, that is to conduct a comprehensive evaluation of the DL models by accounting for a wide range of scenarios in the presence of both epistemic and aleatoric uncertainty.
Still, in the field of fault diagnosis other datasets have been collected, such as the Paderborn University dataset and the University of Ottawa rolling-element dataset \cite{lessmeier2016condition,sehri2023university}.
Future research developments will be, therefore, devoted to extend the comparison on these datasets, in order to corroborate the evidences derived from the current study, while also investigating the effect of factors that are specific to other domains.

Beside this, future works will be addressed to deepen the analysis in the aleatoric uncertainty domain by including a preliminary stage where a classification task is performed to distinguish among different types of noise. 
This would be useful for overcoming a limitation of the present study, in which the type of noise affecting the data is assumed to be known.
Still with reference to aleatoric uncertainty, it would be also useful to release the assumption of observing only one type of noise at a time and study the joint effect of multiple noise types. 
This could help bridging, further, the potential gap between the present research and real-world applications.
Another interesting study could, finally, consider an active learning approach where domain experts are directly involved in the process for better detection of actual OOD data from the predicted untrustworthy ones.

\section*{Acknowledgment}

The present study has been developed within the HumanTech Project, which is financed by the Italian Ministry of University and Research (MUR) for the 2023-2027 period as part of the ministerial initiative “Departments of Excellence” (L. 232/2016).

The present manuscript as been published as a pre-print~\citep{jalayer2024evaluating}.

\bibliographystyle{abbrvnat}
\bibliography{biblio}

\appendix

\section{Hyperparameters}
\label{app:A}
The hyperparameters of each architecture are listed in the tables below.
\clearpage

\begin{table}[H]
\setlength{\tabcolsep}{3pt}
\renewcommand{\arraystretch}{1.2}
\centering
\resizebox{0.7\linewidth}{!}{
\begin{tabular}{|l|l|l|}
\hline
\textbf{Layer} & \textbf{Parameter} & \textbf{Value} \\ \hline
Input layer & Number of nodes & 512 \\ \hline

\multirow{7}{*}{1st Conv + Pool} 
   & Activation function (Convolutional layer) & ReLU \\
   & Number of filters (Convolutional layer) & 16 \\
   & Kernel size (Convolutional layer) & 64 \\
   & Number of strides (Convolutional layer) & 16 \\
   & Padding type (Convolutional layer) & same \\
   & Batch Normalization (Convolutional layer) & Active \\
   & Number of strides (Pooling layer) & 2 \\ \hline

Dropout layer & Dropout rate & 0.2 \\ \hline

\multirow{7}{*}{2nd Conv + Pool} 
   & Activation function (Convolutional layer) & ReLU \\
   & Number of filters (Convolutional layer) & 32 \\
   & Kernel size (Convolutional layer) & 3 \\
   & Number of strides (Convolutional layer) & 1 \\
   & Padding type (Convolutional layer) & same \\
   & Batch Normalization (Convolutional layer) & Active \\
   & Number of strides (Pooling layer) & 2 \\ \hline

Dropout layer & Dropout rate & 0.2 \\ \hline
LSTM & Number of nodes & 64 \\ \hline

\multirow{2}{*}{Dense layer} 
   & Number of nodes & 100 \\
   & Activation function & Sigmoid \\ \hline

Dropout layer & Dropout rate & 0.2 \\ \hline

\multirow{2}{*}{Output layer} 
   & Number of nodes & 6 \\
   & Activation function & Softmax \\ \hline
\end{tabular}
}
\caption{\textbf{Hyperparameters of the ConvLSTM-D architecture.}}
\label{tab:A1}
\end{table}

\begin{table}[H]
\setlength{\tabcolsep}{3pt}
\renewcommand{\arraystretch}{1.2}
\centering
\resizebox{0.7\linewidth}{!}{
\begin{tabular}{|l|l|l|}
\hline
\textbf{Layer} & \textbf{Parameter} & \textbf{Value} \\ \hline
Input layer & Number of nodes & 512 \\ \hline

\multirow{7}{*}{1st Conv + Pool} 
   & Activation function (Convolutional layer) & ReLU \\
   & Number of filters (Convolutional layer) & 16 \\
   & Kernel size (Convolutional layer) & 64 \\
   & Number of strides (Convolutional layer) & 16 \\
   & Padding type (Convolutional layer) & same \\
   & Batch Normalization (Convolutional layer) & Active \\
   & Number of strides (Pooling layer) & 2 \\ \hline

\multirow{7}{*}{2nd Conv + Pool} 
   & Activation function (Convolutional layer) & ReLU \\
   & Number of filters (Convolutional layer) & 32 \\
   & Kernel size (Convolutional layer) & 3 \\
   & Number of strides (Convolutional layer) & 1 \\
   & Padding type (Convolutional layer) & same \\
   & Batch Normalization (Convolutional layer) & Active \\
   & Number of strides (Pooling layer) & 2 \\ \hline

LSTM & Number of nodes & 64 \\ \hline

\multirow{2}{*}{\begin{tabular}[c]{@{}l@{}} Bayesian  Dense \\ layer \end{tabular}} 
   & Number of nodes & 100 \\
   & Activation function & Sigmoid \\ \hline

\multirow{2}{*}{Output layer} 
   & Number of nodes & 6 \\
   & Activation function & Softmax \\ \hline
\end{tabular}
}
\caption{\textbf{Hyperparameters of the BNN architecture.}}
\label{tab:A2}
\end{table}

\begin{table}[H]
\setlength{\tabcolsep}{3pt}
\renewcommand{\arraystretch}{1.2}
\centering
\resizebox{0.7\linewidth}{!}{
\begin{tabular}{|l|l|l|l|}
\hline
\multirow{12}{*}{\begin{tabular}[c]{@{}c@{}} \textbf{For all} \\ \textbf{Base} \\ \textbf{Learners} \end{tabular}} 
   & \textbf{Layer} & \textbf{Parameter} & \textbf{Value} \\ \cline{2-4} 

   & \multirow{7}{*}{Conv + Pool}  
     & Activation function (Convolutional layer) & ReLU \\
   & & Number of filters (Convolutional layer) & 16 \\
   & & Kernel size (Convolutional layer) & 3 \\
   & & Number of strides (Convolutional layer) & 1 \\
   & & Padding type (Convolutional layer) & same \\
   & & Batch Normalization (Convolutional layer) & Active \\
   & & Number of strides (Pooling layer) & 2 \\ \cline{2-4} 

   & \multirow{2}{*}{Dense layer}  
     & Number of nodes & 64 \\
   & & Activation function & Sigmoid \\ \cline{2-4} 

   & \multirow{2}{*}{Output layer} 
     & Number of nodes & 6 \\
   & & Activation function & Softmax \\ \hline
\end{tabular}
}
\caption{\textbf{Hyperparameters of the De1 architecture.}}
\label{tab:A3}
\end{table}

\begin{table}[H]
\setlength{\tabcolsep}{3pt}
\renewcommand{\arraystretch}{1.2}
\centering
\resizebox{0.7\linewidth}{!}{
\begin{tabular}{|l|l|l|l|}
\hline
\multirow{12}{*}{\begin{tabular}[c]{@{}c@{}} \textbf{For two} \\ \textbf{Base} \\ \textbf{Learners} \end{tabular}} 
   & \textbf{Layer} & \textbf{Parameter} & \textbf{Value} \\ \cline{2-4} 
   & Input layer & Number of nodes & 512 \\ \cline{2-4} 

   & Conv + Pool  
     & Activation function (Convolutional layer) & ReLU \\
   & & Number of filters (Convolutional layer) & 16 \\
   & & Kernel size (Convolutional layer) & 3 \\
   & & Number of strides (Convolutional layer) & 1 \\
   & & Padding type (Convolutional layer) & same \\
   & & Batch Normalization (Convolutional layer) & Active \\
   & & Number of strides (Pooling layer) & 2 \\ \cline{2-4} 

   & Dense layer  
     & Number of nodes & 64 \\
   & & Activation function & Sigmoid \\ \cline{2-4} 

   & Output layer 
     & Number of nodes & 6 \\
   & & Activation function & Softmax \\ \hline

\multirow{20}{*}{\begin{tabular}[c]{@{}c@{}} \textbf{For two} \\ \textbf{Base} \\ \textbf{Learners} \end{tabular}} 
   & Input layer & Number of nodes & 512 \\ \cline{2-4} 

   & 1st Conv + Pool
     & Activation function (Convolutional layer) & ReLU \\
   & & Number of filters (Convolutional layer) & 16 \\
   & & Kernel size (Convolutional layer) & 64 \\
   & & Number of strides (Convolutional layer) & 16 \\
   & & Padding type (Convolutional layer) & same \\
   & & Batch Normalization (Convolutional layer) & Active \\
   & & Number of strides (Pooling layer) & 2 \\ \cline{2-4} 

   & 2nd Conv + Pool
     & Activation function (Convolutional layer) & ReLU \\
   & & Number of filters (Convolutional layer) & 32 \\
   & & Kernel size (Convolutional layer) & 3 \\
   & & Number of strides (Convolutional layer) & 1 \\
   & & Padding type (Convolutional layer) & same \\
   & & Batch Normalization (Convolutional layer) & Active \\
   & & Number of strides (Pooling layer) & 2 \\ \cline{2-4} 

   & LSTM layer & Number of nodes & 64 \\ \cline{2-4} 

   & Dense layer 
     & Number of nodes & 100 \\
   & & Activation function & Sigmoid \\ \cline{2-4} 

   & Output layer 
     & Number of nodes & 6 \\
   & & Activation function & Softmax \\ \hline
\end{tabular}
}
\caption{\textbf{Hyperparameters of the De2 architecture.}}
\label{tab:A4}
\end{table}

\section{Entropy distribution of ID and OOD data in aleatoric uncertainty}
\label{app:B}
The entropy distribution of ID and OOD data for aleatoric uncertainty for noisy data by Impulse, Rayleigh, and Gaussian noise types are represented here.
 
 \begin{figure}[H]
 \centering
 \includegraphics[width=\linewidth]{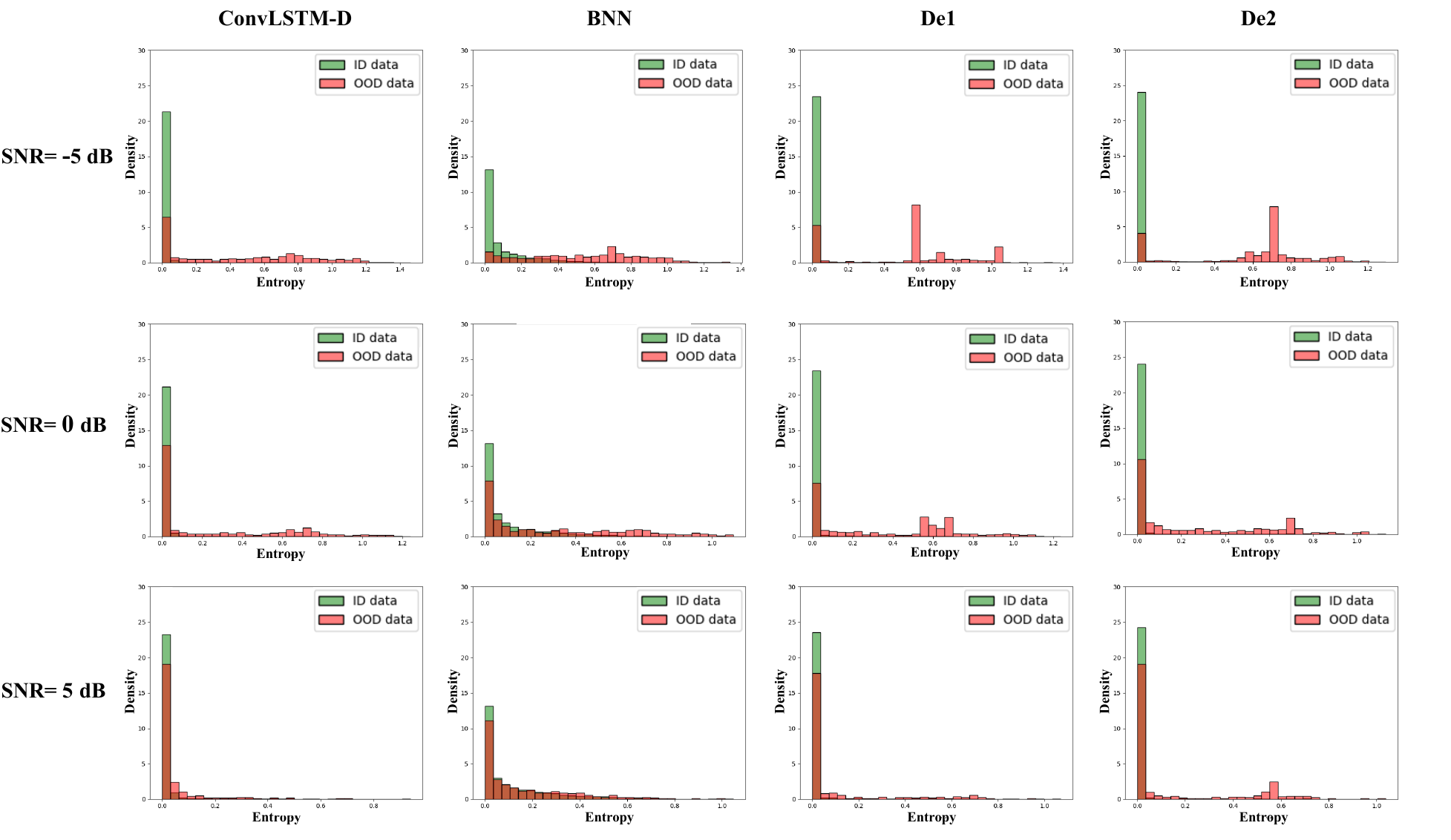}
 \caption{\textbf{The entropy distribution of ID (green) and OOD (red) data is represented. The noise type is Impulse. The top, middle, and bottom sections correspond to different noise levels: SNR = -5 dB, 0 dB, and 5 dB, respectively. From left to right, the figures represent the four architectures: ConvLSTM-D, BNN, De1, and De2.}}
 \label{fig:B1}
\end{figure}

 \begin{figure}[H]
 \centering
 \includegraphics[width=\linewidth]{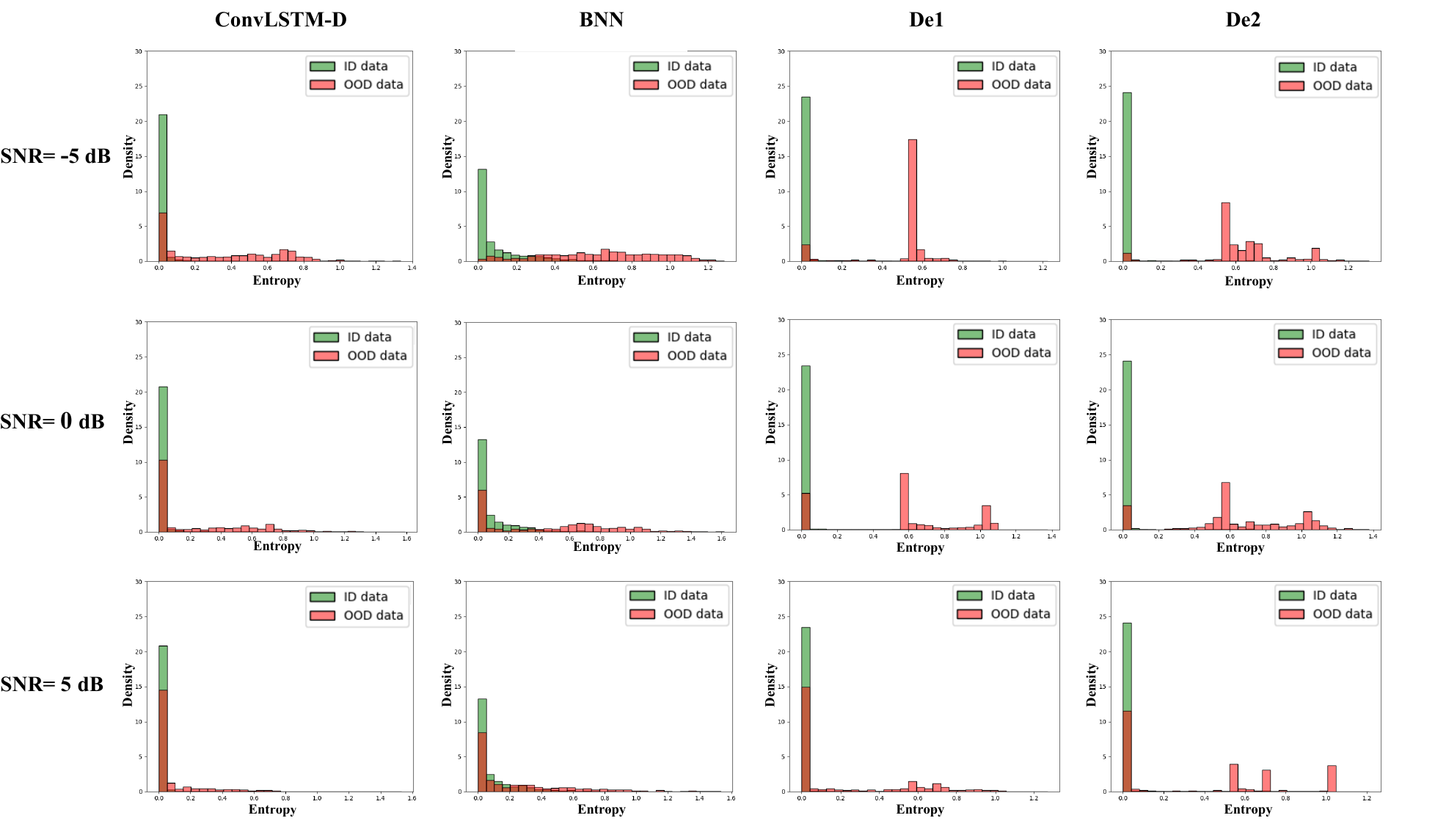}
 \caption{\textbf{The entropy distribution of ID (green) and OOD (red) data is represented. The noise type is Rayleigh. The top, middle, and bottom sections correspond to different noise levels: SNR = -5 dB, 0 dB, and 5 dB, respectively. From left to right, the figures represent the four architectures: ConvLSTM-D, BNN, De1, and De2.}}
 \label{fig:B2}
\end{figure}

 \begin{figure}[H]
 \centering
 \includegraphics[width=\linewidth]{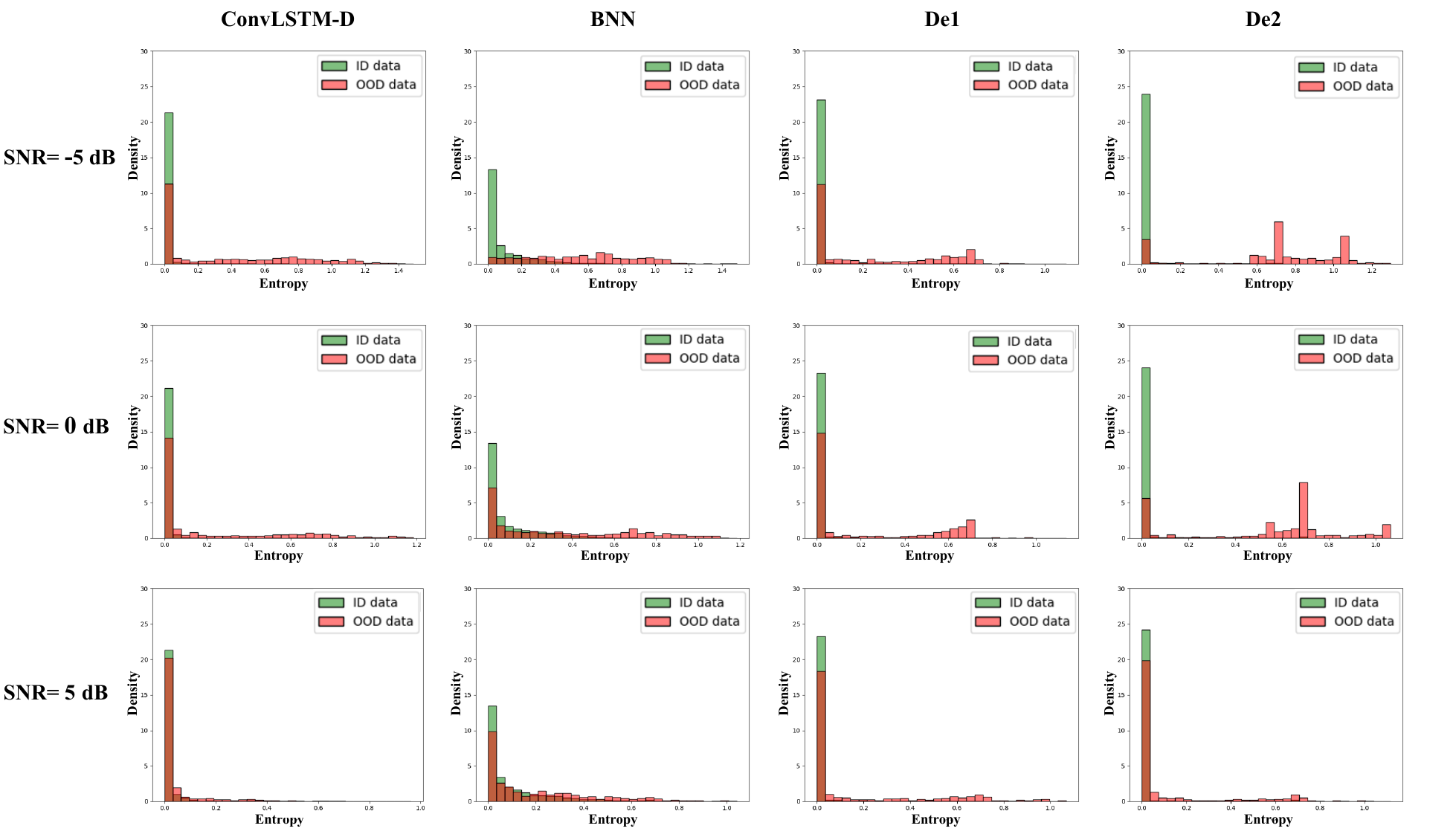}
 \caption{\textbf{The entropy distribution of ID (green) and OOD (red) data is represented. The noise type is Gaussian. The top, middle, and bottom sections correspond to different noise levels: SNR = -5 dB, 0 dB, and 5 dB, respectively. From left to right, the figures represent the four architectures: ConvLSTM-D, BNN, De1, and De2.}}
 \label{fig:B3}
\end{figure}

\end{document}